\documentclass[10pt,journal,cspaper,compsoc]{IEEEtran}

\usepackage{graphicx}
\usepackage[cmex10]{amsmath}
\usepackage{amssymb}

\usepackage{tikz}

\usepackage{comment}
\usepackage{mathtools}
\usepackage{pgfplots}
\usetikzlibrary{matrix}
\usepgfplotslibrary{groupplots}
\pgfplotsset{compat=newest}
\usepackage{pgfplotstable}

\ifCLASSOPTIONcompsoc
  \usepackage[nocompress,noadjust]{cite}
\else
  \usepackage[noadjust]{cite}
\fi
\ifCLASSINFOpdf
\else
\fi
\usepackage[caption=false,font=footnotesize,labelfont=sf,textfont=sf]{subfig}
\begin{document}
%
\title{Approximated and User Steerable tSNE for\\ Progressive Visual Analytics}
%
%
%
%
%

\author{Nicola~Pezzotti,~Boudewijn~P.F.~Lelieveldt,~Laurens~van~der~Maaten,\\~Thomas~H\"ollt,~Elmar~Eisemann,~and~Anna~Vilanova

\IEEEcompsocitemizethanks{\IEEEcompsocthanksitem N.~Pezzotti,~B.~P.F.~Lelieveldt,~L.~van~der~Maaten,~T.~H\"ollt,~E.~Eisemann,~and~A.~Vilanova are with the Department of Intelligent Systems, Delft University of Technology, Delft, the Netherlands.\protect\\
\IEEEcompsocthanksitem B.~P.F.~Lelieveldt is with the Division of Image Processing, Department of Radiology, Leiden University Medical Center, Leiden, the Netherlands.}
\thanks{Manuscript received August 4, 2015; revised -, -.}}

%
%

\markboth{IEEE TRANSACTIONS ON VISUALIZATION AND COMPUTER GRAPHICS, VOL. -, NO. -, MONTH -}%
{Shell \MakeLowercase{\textit{et al.}}: }
%



\IEEEtitleabstractindextext{%
\begin{abstract}
Progressive Visual Analytics aims at improving the interactivity in existing analytics techniques by means of visualization as well as interaction with intermediate results.
One key method for data analysis is dimensionality reduction, for example, to produce 2D embeddings that can be visualized and analyzed efficiently. t-Distributed Stochastic Neighbor Embedding (tSNE) is a well-suited technique for the visualization of several high-dimensional data. tSNE can create meaningful intermediate results but suffers from a slow initialization that constrains its application in Progressive Visual Analytics. We introduce a controllable tSNE approximation (A-tSNE), which trades \emph{•}off speed and accuracy, to enable interactive data exploration. We offer real-time visualization techniques, including a density-based solution and a Magic Lens to inspect the degree of approximation. With this feedback, the user can decide on local refinements and steer the approximation level during the analysis. We demonstrate our technique with several datasets, in a real-world research scenario and for the real-time analysis of high-dimensional streams to illustrate its effectiveness for interactive data analysis. 
\end{abstract}

\begin{IEEEkeywords}
High Dimensional Data, Dimensionality Reduction, Progressive Visual Analytics, Approximate Computation
\end{IEEEkeywords}}

\maketitle

\IEEEdisplaynontitleabstractindextext

%
\IEEEpeerreviewmaketitle

\ifCLASSOPTIONcompsoc
\IEEEraisesectionheading{\section{Introduction}\label{sec:introduction}}
\else
\section{Introduction}
\label{sec:introduction}
\fi

\textcolor{red}{\textbf{IMPORTANT: this work has been extended and the final version is published on the \textit{Transaction on Visualization and Computer Graphics} journal. Please refer to it using the following DOI: http://dx.doi.org/10.1109/TVCG.2016.2570755}}\\
\\

\IEEEPARstart{V}{isual} analysis of high dimensional data is a challenging process.
Direct visualizations such as parallel coordinates~\cite{ParallelCoordinates} or scatterplot matrices~\cite{SPLOM} work well for a few dimensions, but do not scale to hundreds or thousands of dimensions.
Typically indirect visualization is used for these cases.
First the dimensionality of the data is reduced, usually to two or three dimensions, then the remaining dimensions are used to lay out the data for visual inspection, for example in a two dimensional scatterplot.
Dimensionality reduction techniques have been an active field of research in the last years, resulting in a number of viable techniques~\cite{Maaten08dimensionalityreduction}.
A variant of tSNE~\cite{LastFromLaurens,ref:tsne}, the Barnes Hut SNE~\cite{ref:bhtsne} has been accepted as the state of the art for non-linear dimensionality reduction applied to visual analysis of high-dimensional space in several application areas, such as life sciences~\cite{visne,babcock2013deorphanizing,becher2014high,laczny2014alignment,mahfouz2014visualizing,ACCENSE}.
tSNE produces 2D and 3D embeddings that are meant to preserve local structure in the high-dimensional data.
The analyst looks at the embeddings with the goal to identify clusters or patterns that are used to generate new hypothesis on the data, however, the computational complexity of this technique does not allow direct employment in interactive systems.
This limitation makes the analytic process a time consuming task that can take hours, or even days, to adjust the parameters and generate the right embedding to be analyzed.

Recently Stolper et al.~\cite{progressiveVisualAnalytics}, as well as M\"uhlbacher et al.~\cite{OpeningBlackBox} introduced Progressive Visual Analytics.
The idea of Progressive Visual Analytics is to provide the user with meaningful intermediate results, in case computation of the final result is too costly.
Based on these intermediate results the user can start with the analysis process.
M\"uhlbacher et al. also provide a set of requirements, which an algorithm needs to fulfill in order to be suitable for Progressive Visual Analytics.
Based on these requirements they analyze a series of different algorithms, commonly deployed in visual analytics systems, and conclude that, for example, tSNE fulfills all requirements.
The reason being that the minimization in tSNE is built on the iterative gradient descent technique \cite{ref:tsne} and can therefore be used directly for a per-iteration visualization, as well as interaction with the intermediate results.
However, M\"uhlbacher et al. ignore the fact that the distances in the high-dimensional space need to be precomputed to start the minimization process.
In fact this initialization process is dominating the overall performance of tSNE.
Even with a per-iteration visualization of the intermediate results~\cite{pive,OpeningBlackBox,progressiveVisualAnalytics} the initialization time will force the user to wait minutes, or even hours, before the first intermediate result can be generated on a state-of-the-art desktop computer.
Every modification of the data, for example, the addition of data-points or a change in the high-dimensional space, will force the user to wait for the full reinitialization of the algorithm.

In this work we present A-tSNE, a novel approach to adapt the complete tSNE pipeline, including a distance computation for the Progressive Visual Analytics paradigm.
Instead of precomputing precise distances we propose to approximate the distances using Approximated K-Nearest Neighborhood queries.
This allows to start the computation of the iterative minimization nearly instantly after loading the data.
Based on the intermediate results of the tSNE, the user can now start the interpretation process of the data immediately. 
Further, we modified the gradient descent of tSNE such that it allows the incorporation of updated data during the iterative process.
This allows for a continuous refining of approximated neighborhoods in the background and an update of the embedding without restarting the optimization and, eventually, arriving at the precise solution.
Furthermore, we allow the user to steer the level of approximation by selecting points of interest, such as clusters which appear in the very early stages of the optimization and enable an interactive exploration of the high-dimensional data.


Our contributions are as follows:
\begin{enumerate}
\item We present A-tSNE, a twofold evolution of the tSNE algorithm, which
\begin{enumerate}
\item minimizes initialization time and as such enables immediate inspection of preliminary computation results.
\item allows for interactive modification, removal or addition of high-dimensional data, without disrupting the visual analysis process.
\end{enumerate}
\item Using a set of standard benchmark data sets, we show large performance increases of A-tSNE compared to the state of the art while maintaining high precision.
\item We developed an interactive system for a visual analysis of high dimensional data, allowing the user to inspect and steer the level of approximation and we illustrate the benefits of exploratory possibilities in a real-world research scenario and for the real-time analysis of high-dimensional streams.
\end{enumerate}

\section{Related work}
\label{sec:related_work}
The tSNE~\cite{ref:tsne} algorithm builds the foundation of this work.
As described above, tSNE is used for visualization of high-dimensional data in a wide field of applications, from life sciences to the analysis of deep-learning algorithms~\cite{visne,babcock2013deorphanizing,becher2014high,frome2013devise,decaf,laczny2014alignment,mahfouz2014visualizing,AINature,ACCENSE}. 
tSNE is a non-linear dimensionality-reduction algorithm that aims at preserving local structures in the embedding, whilst showing global information, such as the presence of clusters at several scales.  

tSNE's computational and memory complexity is $O(N^2)$, where $N$ is the number of data-points, which constrains the application of the technique. An evolution of the algorithm, called Barnes-Hut-SNE (BH-SNE)~\cite{ref:bhtsne,LastFromLaurens}, reduces the computational complexity to $O(N \log(N) )$ and the memory complexity to $O(N)$. 
This approach was also developed in parallel by Yang et al.~\cite{ref:bhsne_2}.
However, despite the increased performance, it still cannot be used to interactively explore the data in a desktop environment.

Interactive performance is at the center of the latest developments in Visual Analytics. New analytical tools and algorithms, which are able to trade accuracy for speed and offer the possibility to interactively refine results \cite{FeketeOBB,TrustMe}, are needed to deal with the scalability issues of existing analytics algorithms like tSNE. 
M\"uhlbacher et al.~\cite{OpeningBlackBox} defined different strategies to increase the user involvement in existing algorithms. 
They provide an in-depth analysis on how the interconnection between the visualization and the analytic modules can be achieved.
Stolper et al.~\cite{progressiveVisualAnalytics} defined the term \emph{Progressive Visual Analytics} that describes techniques that allow the analyst to directly interact with the analytics process. 
Visualization of intermediate results is used to help the user, for example, to find optimal parameter settings or filter the data.
The authors also provide a guideline for the design of Progressive Visual Analytics. 
Many algorithms, however, are not suited right away for Progressive Visual Analytics since the production of intermediate results is computationally too intensive or they do not generate useful intermediate results at all.
tSNE is an example of such an algorithm because of its initialization process.

To overcome this problem, we propose to compute an approximation of tSNE's initialization stage, followed by a user steerable~\cite{mulder1999survey} refinement of the level of approximation. 
To compute the conditional probabilities needed by BH-SNE, a K-Nearest Neighborhood~(KNN) search must be evaluated for each point in the high-dimensional space. 
Under these conditions, a traditional algorithm and data structure, such as a KD-Tree~\cite{kdtree}, will not perform well. 
In the BH-SNE~\cite{ref:bhtsne} algorithm, a Vantage-Point Tree~\cite{vptree} is used for the KNN search, but it is slow to query.
In this work, we propose to use an approximated computation of the KNN in the initialization stage to start the analysis as soon as possible.
The level of approximation is then refined on the fly during the analytics process.

Other dimensionality-reduction algorithms implement approximation and steerability to increase performance as well.
For example MDSteer~\cite{SteerableMDS} works on a subset of the data and allows the user to control the insertion of points by selecting areas in the reduced space. 
Yang et al.~\cite{yang2006fast} present a dimensionality-reduction technique using a dissimilarity matrix as input.
By means of a divide-and-conquer approach, the computational complexity of the algorithm can be reduced. 
Multiple other techniques provide steerability by means of guiding the dimensionality reduction via user input.
For example Joja et al.~\cite{P0} and Paulovich et al.~\cite{P2} let the user place a small number of control points. 
In other work, Paulovich et al.~\cite{P1}, propose the use of a non-linear dimensionality-reduction algorithm on a small number of automatically-selected control points.
For these techniques the position of the data points is then finally obtained by linear-interpolation schemes that make use of the control points.
All these techniques, however, limit the non-linear dimensionality reduction to a subset of the dataset.
In this work, we provide a way to directly use the complete data allowing the analyst to immediately start the analysis on all the data points.

Ingram and Munzner's Q-SNE~\cite{qsne} is based on a similar idea, also using Approximated KNN queries for the computation of the high-dimensional similarities.
However, their work is designed to exploit the sparse structure of high-dimensional spaces obtained from document collections and, therefore, cannot be used for the analysis of dense high-dimensional spaces as presented in this work.
Furthermore, we provide a visualization of the degree of approximation, the ability to steer the computation by refining the approximation level based on user input and the ability to interactively manipulate the high-dimensional data.

Density-based visualization of the tSNE embedding has been used in several works~\cite{visne,ACCENSE,LastFromLaurens}, however, they employ slow to compute offline techniques.
In our work, we integrate real time Kernel Density Estimation (KDE) as described by Lampe and Hauser~\cite{ref:KDE_Scatter_plots}.
The interaction with the embedding is important to allow the analyst to explore the high-dimensional data. 
Selection operations in the embedding and the visualization of the data in a coordinated multiple-view system are necessary to enable this exploration. 
The iVisClassifier system~\cite{ivisclassifier} is an example of such a solution. 
In our work, we take a similar approach, providing a coordinated multiple-view framework for the visualization of a selection in the embedding.

\section{\MakeLowercase{t}SNE}
\label{sec:tSNE}
In this section, we provide a short introduction to tSNE~\cite{ref:tsne}, which is necessary to explain our contribution.
tSNE interprets the overall distances between data-points in the high-dimensional space as a symmetric joint-probability distribution $P$.
Likewise a joint-probability distribution $Q$ is computed, that describes the similarity in the low-dimensional space.
The goal is to achieve a representation, referred to as \emph{embedding}, in the low dimensional space where $Q$ faithfully represents $P$.
This is achieved by optimizing the positions in the low-dimensional space to minimize the cost function $C$ given by the Kullback-Leibler ($KL$) divergence between the joint-probability distributions $P$ and $Q$:

\begin{equation}\label{eq:KL}
 C(P,Q) = KL(P||Q) = \sum_{i=1}^N\sum_{j=1}^N p_{ij} \ln\left(\frac{p_{ij}}{q_{ij}}\right)
\end{equation}

Given two data points $\mathbf{x}_i$ and $\mathbf{x}_j$ the probability $p_{ij}$ models the similarity of these points in the high-dimensional space. To this extent, for each point a Gaussian kernel, $P_i$, is chosen whose variance $\sigma_i$ is defined according to the local density in the high-dimensional space and then $p_{ij}$ is described as follows: 

\begin{equation} \label{eq:tsne_HD}
p_{ij} = \frac{p_{i|j} + p_{j|i}}{2N},
\end{equation} 

\begin{equation} \label{eq:P_j|i}
\text{where} \quad  p_{j|i}=\frac{exp(-(||\mathbf{x}_i-\mathbf{x}_j||^2)/(2 \sigma_i^2))}{\sum_{k\ne i}^N exp(-(||\mathbf{x}_i-\mathbf{x}_k||^2)/(2 \sigma_i^2))}
\end{equation}

$p_{j|i}$ can be seen as a relative measure of similarity based on the local neighborhood of a data-point $\mathbf{x}_i$. The perplexity value $\mu$ is a user-defined parameter that describes the effective number of neighbors considered for each data-point. The value of $\sigma_i$ is chosen such that for fixed $\mu$ and each $i$:

\begin{equation} \label{eq:perplexity}
\mu = 2^{-\sum_{j}^N p_{j|i} \log_2 p_{j|i}}
\end{equation}

A \emph{Student's t-Distribution} with one degree of freedom is used to compute the joint-probability distribution in the low-dimensional space $Q$, where the positions of the data-points should be optimized. Given two low-dimensional points $\mathbf{y}_i$ and $\mathbf{y}_j$, the probability $q_{ij}$ that describes their similarity is given by:

\begin{equation} \label{eq:Q}
q_{ij} = \left((1+||\mathbf{y}_i - \mathbf{y}_j||^2) Z \right)^{-1}
\end{equation}

\begin{equation} \label{eq:Z_normalization_factor}
\text{with} \quad Z = \sum_{k=1}^N\sum_{l \ne k}^N(1+||\mathbf{y}_k-\mathbf{y}_l||^2)^{-1}
\end{equation}

The gradient of the Kullback-Leibler divergence between $P$ and $Q$ is used to minimize $C$ (see Eq.~\ref{eq:KL}). It indicates the change in position of the low-dimensional points for each step of the gradient descent and is given by:

\begin{align} \label{eq:C_gradient_n_body}
\frac{\delta C}{\delta \mathbf{y}_i} &= 4 \sum_{i=1}^N( F^{\text{attr}}_i-F^{\text{rep}}_i)\\
&=4 \sum_{i=1}^N( \sum_{j \ne i}^N p_{ij}q_{ij}Z(\mathbf{y}_i-\mathbf{y}_j) - \sum_{j \ne i}^N q_{ij}^2Z(\mathbf{y}_i-\mathbf{y}_j))
\end{align}

The gradient descent can be seen as a \emph{N-body simulation}~\cite{NBodySimulation}, where each data-point exerts an attractive and a repulsive force on all the other points ($F^{\text{attr}}_i$ and $F^{\text{rep}}_i$).

\subsection{Barnes-Hut-SNE}
\label{sec:Barnes-Hut-SNE}
In the original tSNE, the force is computed using a brute-force approach, resulting in computational and memory complexity of $O(N^2)$.
Barnes-Hut-SNE (BH-SNE)~\cite{ref:bhtsne,LastFromLaurens} is an evolution of the tSNE algorithm that introduces two different approximations to reduce the computational complexity to $O(N\log(N))$ and the memory complexity to $O(N)$.

The first approximation is based on the observation that the probability $p_{ij}$ is infinitesimal if $\mathbf{x}_i$ and $\mathbf{x}_j$ are dissimilar. 
Therefore, the similarities of a data-point $\mathbf{x}_i$ can be computed taking into account only the points that belong to the set of nearest neighbors $\mathcal{N}_i$. 
The cardinality of $\mathcal{N}_i$ can be set to $K = \lfloor 3\mu\rfloor$, where $\mu$ is the user-selected perplexity and $\lfloor \cdot \rfloor$ describes a rounding to the next-lower integer. 
Without compromising the quality of the embedding~\cite{ref:bhtsne,LastFromLaurens}, we can adopt a sparse approximation of the high-dimensional similarities. Eq.~\ref{eq:P_j|i} can now be written as follows:

\begin{equation} \label{eq:BH_P_j|i}
p_{j|i} =
  \begin{cases}
   \frac{exp(-(||\mathbf{x}_i-\mathbf{x}_j||^2)/(2 \sigma_i^2))}{\sum^N_{k\ne i} exp(-(||\mathbf{x}_i-\mathbf{x}_k||^2)/(2 \sigma_i^2))} & \text{if } j \in \mathcal{N}_i \\
   0       & \text{otherwise}
  \end{cases}
\end{equation}

The computation of the K-Nearest Neighbors is performed using a Vantage-Point Tree (VP-Tree)~\cite{vptree}. 
A VP-Tree is data structure that computes KNN queries in a high-dimensional metric space, in $O(\log(N))$ time. 
It is a binary tree that stores for each non leaf-node a hyper-sphere centered on a data-point. 
The left children of each node will contain the points that reside inside the hyper-sphere, whereas the right one will contain the points outside it.

The second approximation makes use of the formulation of the gradient presented in Eq.~\ref{eq:C_gradient_n_body}. 
As described above tSNE can be seen as an N-body simulation and thus the Barnes-Hut algorithm~\cite{barneshut} can be used to reduce the computational complexity to $O(N \log(N))$. 
For further details, we refer to van der Maaten~\cite{ref:bhtsne,LastFromLaurens}. 

\section{A-tSNE for Progressive \\Visual Analytics}

\begin{figure}[t]
  \centering
    \subfloat[Progressive Visual Analytics workflow for tSNE.]{
      \includegraphics[width = 0.9 \linewidth]{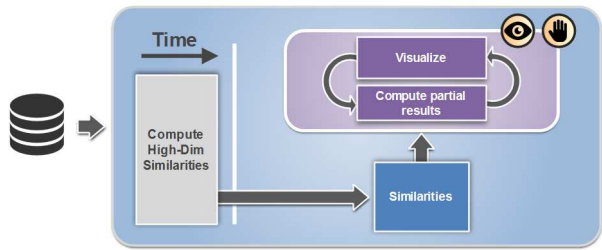}
      \label{fig:wokflows_old}
    }\\
    \subfloat[Progressive Visual Analytics workflow for A-tSNE.]{
      \includegraphics[width = 0.9 \linewidth]{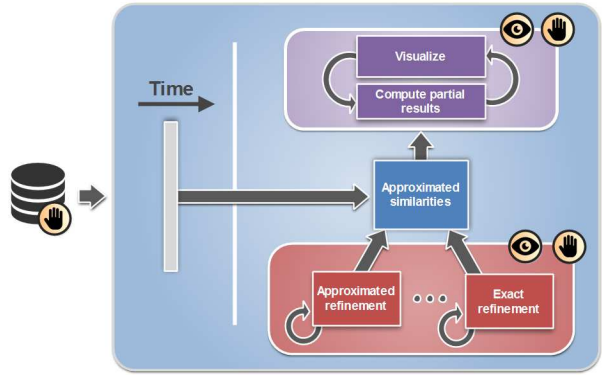}
      \label{fig:wokflow}
    }
  \caption{\textbf{Comparison between the traditional and our tSNE workflow.} The eye icon marks modules which produce output for visualization, whereas the hand icon marks modules that allow manipulation by the user. The increased performance of the similarity computation allows the user to seamlessly manipulate the input data. The level of approximation can be visualized and the user can steer the refinement process to interesting regions.}
  \label{fig:wokflows}
\end{figure}

In this work, we introduce Approximated-tSNE (A-tSNE), an evolution of the BH-SNE algorithm, using approximated computations of high-dimensional similarities to generate meaningful intermediate results. 
The level of approximation can be defined by the user to allow control on the trade off between speed and quality.
The level of approximation can be refined by the analyst in interesting regions of the embedding, making A-tSNE a computational steerable algorithm~\cite{mulder1999survey}.

tSNE is well suited for the application in Progressive Visual Analytics: after the initialization of the algorithm, the intermediate results generated during the iterative optimization process can be interpreted by the analyst while they change over time, as shown in previous work~\cite{pive,OpeningBlackBox}.
Fig.~\ref{fig:wokflows_old} shows a typical Progressive Visual Analytics workflow for tSNE.

Algorithms that can be used in a Progressive Visual Analytics system often have a computational module, e.g. the initialization of the technique,  that cannot be implemented in an iterative way, creating a \emph{speed bump}~\cite{progressiveVisualAnalytics} in the user analysis. 
tSNE is a good example of this limitation, it consists of two computational modules that are serialized. 
In the first part of the algorithm, similarities between high-dimensional points are calculated. 
In the second module, a minimization of the cost function (Eq.~\ref{eq:KL}) is computed by means of a gradient descent. 
The first module, depicted in light grey in Fig.~\ref{fig:wokflows_old}, is slow to compute and does not create any meaningful intermediate results. 
\begin{figure*}[t]
  \centering
  \subfloat[]{
    \includegraphics[width = 0.037 \linewidth]
    {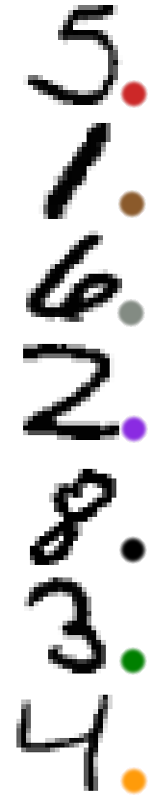}
    \label{fig:mnist_data_points}
  }%
  \subfloat[BH-SNE - Time: 3191.8 s]{
    \includegraphics[width = 0.21 \linewidth]
    {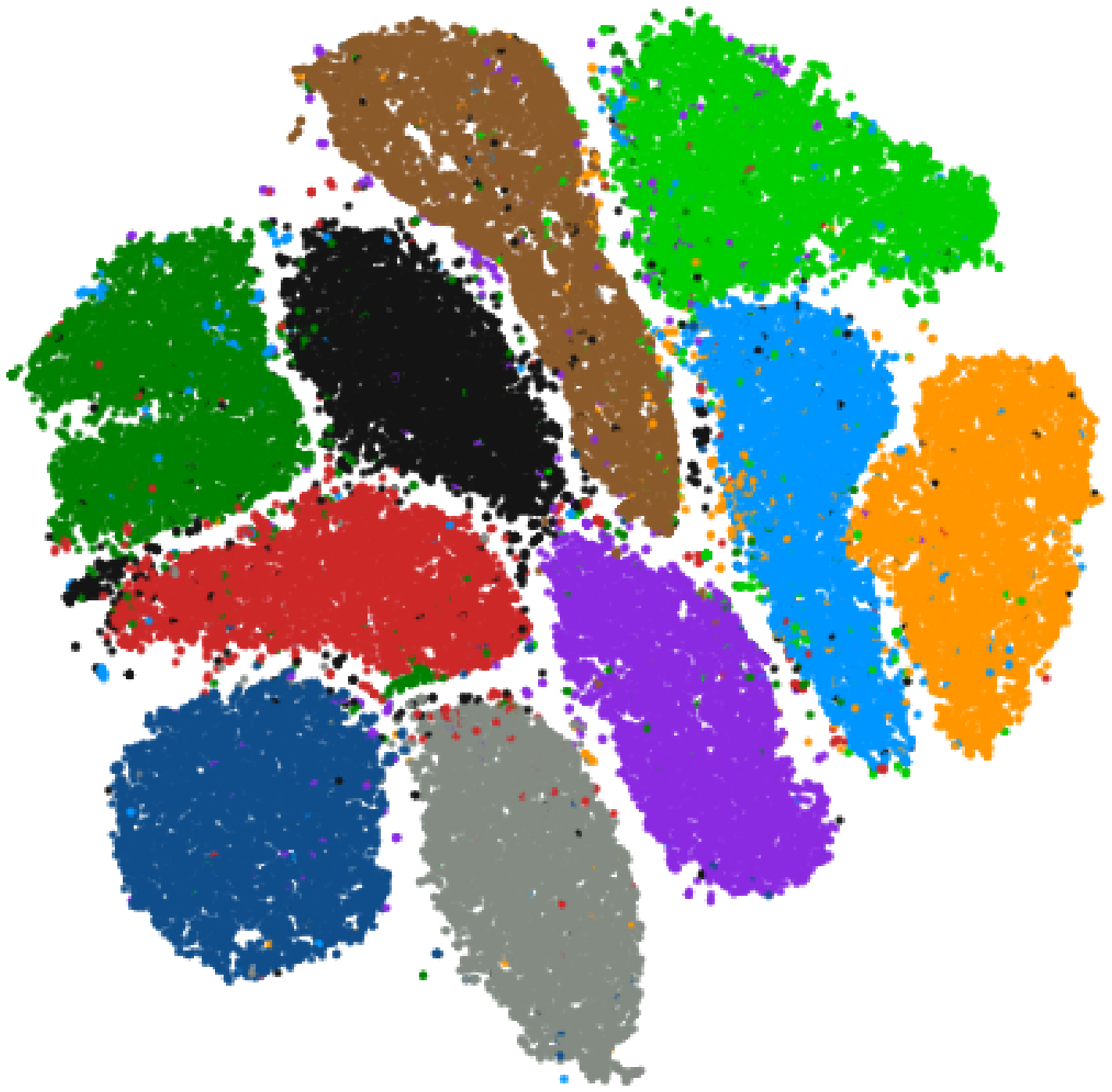}
    \label{fig:embedding_exact}
  }%
  \subfloat[$\rho = 0.34$ - Time: 30.1 s]{
    \includegraphics[width = 0.21 \linewidth]
    {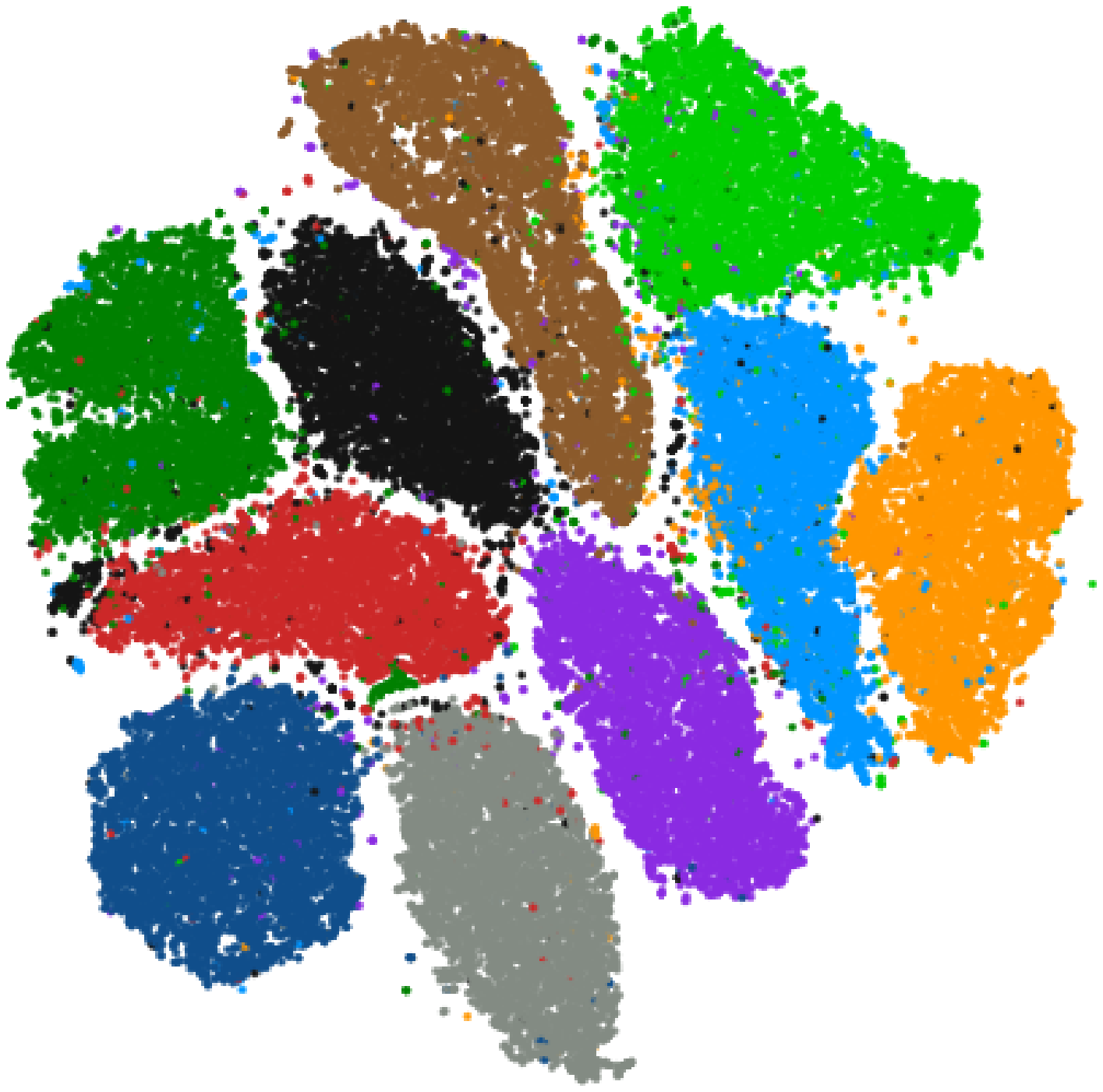}
    \label{fig:embedding_f_4_1024}
  }%
  \subfloat[$\rho = 0.23$ - Time: 20.4 s]{
    \includegraphics[width = 0.21 \linewidth]
    {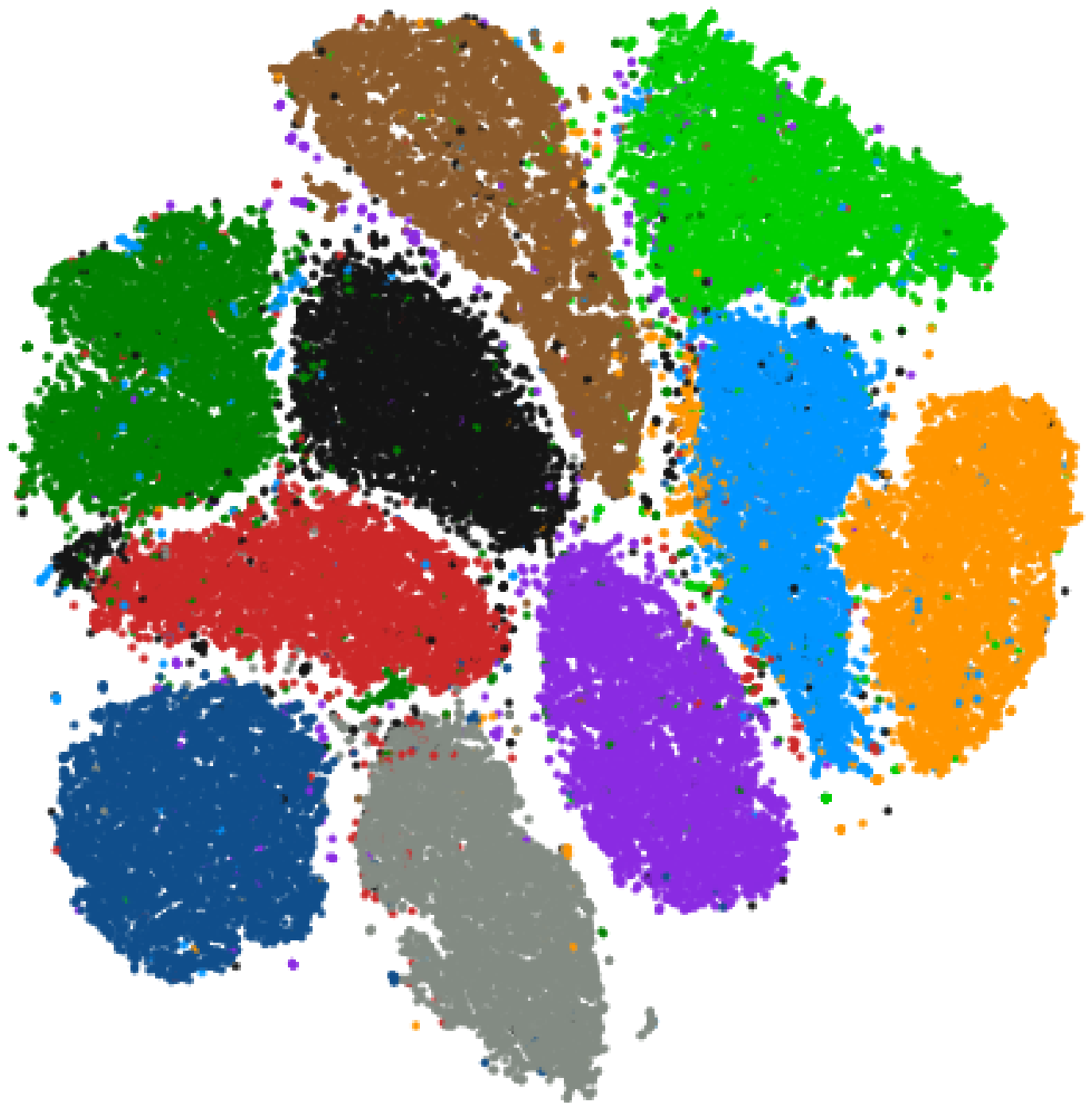}
    \label{fig:embedding_f_2_512}
  }%
  \subfloat[$\rho = 0.07$ - Time: 13.0 s]{
    \includegraphics[width = 0.21 \linewidth]
    {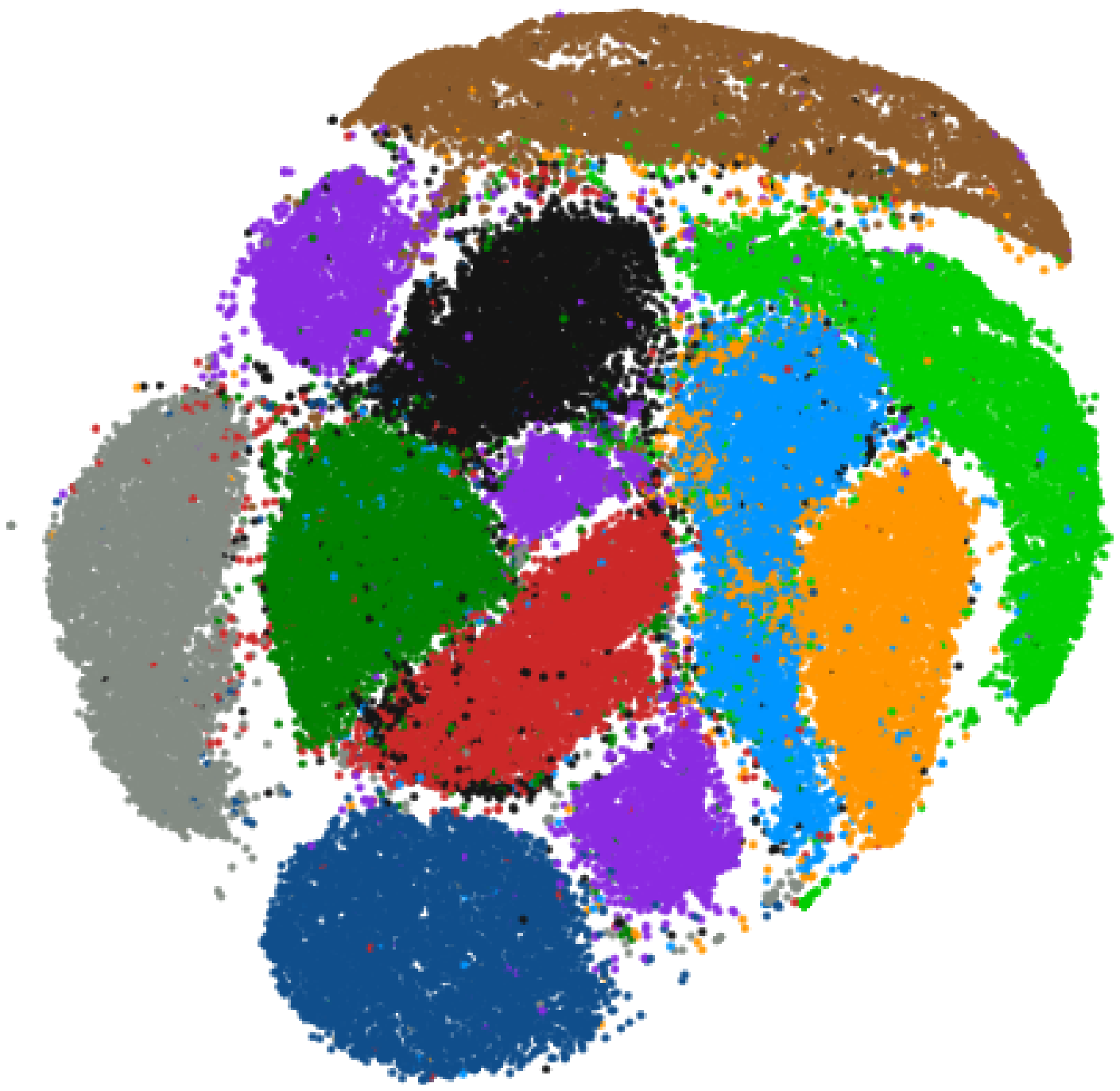}
    \label{fig:embedding_f_1_1}
  }
  \caption{\textbf{Embeddings of the MNIST dataset} using different approximation levels. Each point represents an image of a handwritten digit, few examples are shown in (a). Points are colored according to the classification of the image. 
It can be seen that a reasonable approximation as in (c) and (d) produces nearly identical results, compared to the original BH-SNE (b) two orders of magnitude faster. Even very low precision (e) produces clearly distinguishable clusters, even though the embedding visually differs from (b)-(d).}
  \label{fig:embeddings}
\end{figure*}%

We extend the Progressive Visual Analytics paradigm by introducing approximated computation rather than aiming at exact computations, in the modules that are not suited for a per-iteration visualization. 
Fig.~\ref{fig:wokflow} shows the analytical workflow for A-tSNE.
While the generation and the inspection of the intermediate results is not changed, we introduce a refinement module, depicted in red in Fig.~\ref{fig:wokflow}, which can be used to refine the level of the approximation in the embedding in a concurrent way.
Furthermore, the increased performance of the initialization module and the ability to update the high-dimensional similarities during the gradient descent minimization, allows the analyst to manipulate the high-dimensional data in an interactive way.
We impose the following requirements to our approximation:

\begin{enumerate}
\item The performance gain due to the approximation must be high enough to enable interaction.
\item The approximation should lead to a gradual and controllable degradation of the accurate results.
\item The approximation quality can be measured and visualized to avoid misleading the user.
\item The approximation can be refined during the evolution of the algorithm and, possibly, steered by the user.
\end{enumerate}



In the following section, we describe the A-tSNE algorithm in detail using the MNIST~\cite{MNIST} dataset for illustration.
The dataset consists of 60k labeled gray scale images of handwritten digits (compare Fig.~\ref{fig:embeddings}a).
Each image is represented as a 784 dimensional vector, corresponding to the gray values of the pixels in the image.

\subsection{A-tSNE}
\label{sec:Our_tSNE}

A-tSNE improves the BH-SNE algorithm using fast and Approximated KNN computations to build the approximated high-dimensional joint-probability distribution $P^A$, instead of the exact distribution $P$. 
The cost function $C(P^A,Q^A)$ is then minimized in order to obtain the approximated embedding described by $Q^A$.

The similarity between points can be computed using the set of approximated neighbors $\mathcal{N}^A_i$, instead of the exact neighborhood $\mathcal{N}_i$ (see Eq.~\ref{eq:BH_P_j|i}).
We define the precision of the algorithm as $\rho$.
$\rho$ describes the average percentage of points in the approximated neighborhood $\mathcal{N}^A_i$ that belongs to the exact neighborhood $\mathcal{N}_i$:

\begin{equation} \label{eq:precision}
\rho = \sum_{i=1}^N \frac{\rho_i}{N} \quad \rho_k = \frac{ | \mathcal{N}^A_k \cap \mathcal{N}_k |}{ |\mathcal{N}_k|},
\end{equation}

where $|\cdot|$ indicates the cardinality of the neighborhood.
The cardinality of $\mathcal{N}_k$ is indirectly specified by the user as explained in Sec.~\ref{sec:Barnes-Hut-SNE}, as three times the value of the perplexity parameter $\mu$.
$\rho$ is an input parameter that can be defined by the user, setting $\rho$ to a large value means the joint-probability distribution $P_A$ is similar to $P$, leading to a similar embedding.

To better understand the effect of the approximated queries, it is useful to interpret the BH-SNE algorithm as a force-directed layout algorithm~\cite{fruchterman1991graph}, which acts on an undirected graph created by the KNN relationships. 
A data point $\mathbf{x}_i$ is repelled by all other data-points but to a subset of the data-points given by its neighborhood relationships, where attraction forces are created by a  set of springs which connect $\mathbf{x}_i$ with all the points in $\mathcal{N}_i$.

When specifying a lower precision $\rho$, resulting in a coarser approximation, some springs connecting points, which are close in the high-dimensional space will be missing and instead distant points are connected. 
This will result in a false repulsion between the points missing the connecting spring.
Using $P^A$ reduces the quality of the embedding but improves its computation time by several orders of magnitude.
However, reasonable results can usually be achieved even with low values for $\rho$, because each data point is usually connected to a large number of springs and, therefore, the overall structure can be maintained. 
This observation holds for local as well as global structures. Intuitively, even if two points are no longer connected, they might share a common neighbor, which indirectly connects both.

Fig.~\ref{fig:embeddings} shows the embeddings generated using different precision values $\rho$ for the computation of the high-dimension similarities. 
We use the whole  MNIST dataset as the input and we color each data-point accordingly to the digit it represents for validation purposes. 
Fig.~\ref{fig:embedding_exact} shows the embedding generated with the exact neighborhood, whereas Fig.~\ref{fig:embedding_f_4_1024} shows the embedding generated with a precision of $\rho = 0.34$. It can be seen that similar structures are preserved using approximated neighborhoods. 
Fig.~\ref{fig:embedding_f_1_1} shows the embedding generated with $\rho = 0.07$. 
Even though the embedding visually differs from the exact embedding, depicted in Fig.~\ref{fig:embedding_exact}, the overall clustering of the data is preserved rather well, whilst the time needed for the computation of the similarities is greatly reduced.
Where the original algorithm needs $3191$ seconds for the initialization using a precision of $\rho = 0.34$ we can achieve a speedup of two orders of magnitude, resulting in a computation time of $30$ seconds. 
By using a precision of $\rho = 0.07$, it is further reduced to $13$ seconds.


\subsection{Approximated KNN}
\label{sec:AKKN}

We achieve different levels of precision by means of different parameterizations of an approximated KNN algorithm called \textit{Forest of Randomized Kd-Trees}. 
In this section, we describe this technique and how its parameters can be mapped to the precision $\rho$.

When the dimensionality of the data is high, there are no exact KNN algorithms performing better than linear search~\cite{muja_flann_2009}. 
Therefore, the development of approximated KNN algorithms is needed to deal with high-dimensional spaces. 
A survey on existing algorithms, including an extensive set of experiments can be found in the work of Muja et al.~\cite{muja_flann_2014}. 

In this work, we use a space partitioning technique called \emph{Forest of Randomized KD-Trees}~\cite{forestKDTree} to compute the approximated neighborhoods. 
This technique has proven to be fast and effective in querying of high-dimensional spaces~\cite{muja_flann_2009}. 

A KD-Tree~\cite{kdtree} is a binary tree used to partition a k-dimensional space. 
Each node in the tree is a $k-1$ dimensional hyper-plane, orthogonal to one of the initial k-dimensions, that splits the space into two half spaces. 
The recursive splitting creates a hierarchical partition of the k-dimensional space. 

In a \emph{Forest of Randomized KD-Trees}, a number $\mathcal{T}$ of KD-Trees are generated. 
The splitting hyper-planes are selected by splitting along a randomly selected dimension among the $\mathcal{V}$ dimensions characterized by the highest variance.
A KNN search is computed on all $\mathcal{T}$ KD-Trees, while a maximum number of leaves $\mathcal{L}$ are visited. 
A priority-queue, ordered by increasing distances to the decision boundary, is used to decide which nodes must be visited first across the forest. 
The process is stopped when the necessary number of leaves have been evaluated.

The parameterization of the Forest of Randomized KD-Trees can overburden the typical end user.
To hide this complexity, we integrate the work by Muja et al.~\cite{muja_flann_2009} and expose only the single precision parameter $\rho$ to the user.
The parameters involved in the creation and querying of the \emph{Forest of Randomized KD-Trees}, $(\mathcal{T},\mathcal{V},\mathcal{L})$, are estimated by a minimization process that uses the precision $\rho$ as input, an estimation of the structure of the dataset, e.g., correlation between dimensions, and a desired balance between the building time of the trees and the cost of a single KNN search. For further details we refer to Muja et al.~\cite{muja_flann_2009}.

\subsection{Steerability}
\label{sec:refinement}
A-tSNE is computationally steerable~\cite{mulder1999survey}, in the sense that the user can define the level of approximation to specific, interesting areas.
In this section, we present the changes we made to the BH-SNE algorithm to allow for the refining of the approximation.

The refinement that we propose is done by computing the exact neighborhood for one point at a time. This process leads to a mix of exact and approximated neighborhoods.
For each updated neighborhood, a Gaussian distribution $P_i$ is computed and the sparse joint-probability distribution $P^A$ must be updated accordingly.
This update, however, is not straightforward. First, the symmetrization of $P^A$ in Eq.~\ref{eq:tsne_HD} requires to combine Gaussian distributions enforced by different data-points and, second, the sparse nature of the distribution $P^A$ renders fast updates challenging.

We solve these issues by observing that a direct computation of $P^A$ can be avoided and the distribution can be indirectly obtained using the Gaussian distributions enforced by the K-Nearest Neighbors.
Eq.~\ref{eq:tsne_HD} can be split into two components which correspond only to the Gaussian distributions $P_i$ and $P_j$:


\begin{equation} \label{eq:tsne_HD_refinement}
p_{ij} = \frac{p_{j|i}}{2N} + \frac{p_{i|j}}{2N}
\end{equation} 

Using this formulation, we need to store only one Gaussian distribution for every point, therefore points can be handled individually without any performance loss.
This key point of our technique allows us to refine the high-dimensional similarities in parallel to the gradient descent, and it is the base for the manipulation of the high-dimensional data.
Furthermore, we are not constrained to update the neighborhood of a data-point just once. The analyst can request different levels of approximation for a given area before starting the computation of the exact high-dimensional similarities.
For each data-point we store $\rho_i$ as the requested precision for the neighborhood $\mathcal{N}_i$.


A change in a neighborhood, however, yields a change in the cost function $C$, see Eq.~\ref{eq:KL}, which we are minimizing.
To avoid the risk of getting stuck in a local minimum during the gradient descent, we introduce an optimization strategy called \emph{Selective Exaggeration with Exponential Decay}.

Our strategy is inspired by the optimization strategy called \emph{Early Exaggeration} presented by van der Maaten et al.~\cite{ref:tsne}. 
The idea of \emph{Early Exaggeration} is that, by exaggerating the attractive forces, see Eq.~\ref{eq:C_gradient_n_body}, by a factor $\tau$ during the first $I_\tau$ iterations of the gradient descent, local minima can be avoided. 

Using the \emph{Selective Exaggeration with Exponential Decay}, we apply an exaggeration $\tau$ to the attractive forces acting on a data-point $x_i$ when it is refined. 
The exaggeration is then smoothly removed on a per-point basis using an exponential decay of the exaggeration factor.
This can be interpreted as a localized reinitialization of the gradient descent triggered by user interaction with the embedding.

\subsection{Performance and Accuracy Benchmarking}
\label{sec:results}

In this section, we present a detailed performance analysis of A-tSNE compared to the BH-SNE using several standard benchmark datsets. 
All performance measurements were obtained using a DELL Precision T3600 workstation with a 8-core Intel Xeon E5 1650 CPU @ 3.2GHz, 32GB RAM and a NVIDIA GTX 680.

We apply the same preprocessing steps as presented by van der Maaten~\cite{LastFromLaurens}, without applying a preliminary dimensionality-reduction by means of a Principal Component Analysis.
We use the MNIST dataset~\cite{MNIST} (60k data-points with 784 dimensions), the NORB dataset~\cite{NORB} (24300 data-points with 9216 dimensions), the CIFAR-10 dataset~\cite{CIFAR} (50k points with 1024 dimensions) and the TIMIT dataset~\cite{TIMIT}, consisting of 1M data-points with 39 dimensions. 
Throughout the experiments we used a parameter setup similar to the one used to benchmark the BH-SNE~\cite{ref:bhtsne,LastFromLaurens}. 
We use a fixed perplexity value of $\mu = 30$. 
First, we evaluate the performances of A-tSNE in relation to the parameters $(\mathcal{T},\mathcal{V},\mathcal{L})$ used in the \textit{Forest of Randomized KD-Trees} where, $\mathcal{T}$ is the number of trees generate splitting along the $\mathcal{V}$ dimensions characterized by the highest variance and $\mathcal{L}$ is the number of leaves visited in a single query.
We set the approximation parameters to three different configurations: $\mathcal{T}=4$ $\mathcal{L}=1024$, $\mathcal{T}=2$ $\mathcal{L}=512$ and $\mathcal{T}=1$ $\mathcal{L}=1$, while we fix $\mathcal{V} = 5$ as suggested by Muja et al.~\cite{muja_flann_2009}.

\begin{figure}[b!]
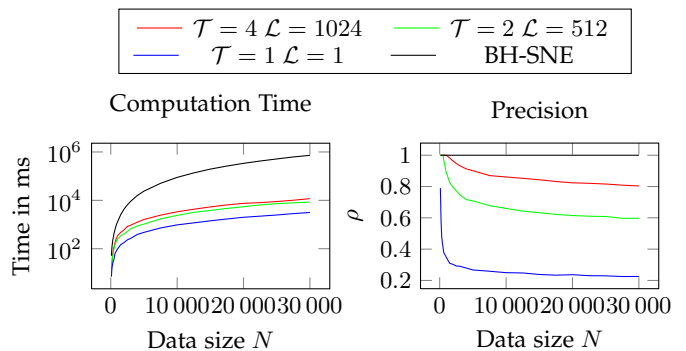

\begin{tikzpicture}

    \begin{groupplot}[group style={group size= 2 by 1,horizontal sep= 1.2cm},width=0.26\textwidth, height=3.5cm]
	    \nextgroupplot[title = {Computation Time}, ylabel={Time in ms},xlabel={Data size $N$},ymode=log,scaled x ticks = false, x tick label style={/pgf/number format/fixed,/pgf/number format/1000 sep = \thinspace},every axis/.append style={font=\small},legend style={font=\tiny}]
			\input{charts/computation_time_MNIST.tex}
                \coordinate (top) at (rel axis cs:0,1);
                \\
                \nextgroupplot[title = {Precision}, ylabel={$\rho$},xlabel={Data size $N$},scaled x ticks = false, x tick label style={/pgf/number format/fixed,/pgf/number format/1000 sep = \thinspace},every axis/.append style={font=\small},legend style={font=\tiny}]
			\input{charts/precision_MNIST.tex}
                \coordinate (bot) at (rel axis cs:1,0);
    \end{groupplot}
\path (top|-current bounding box.north)--
      coordinate(legendpos)
      (bot|-current bounding box.north);
\matrix[
    matrix of nodes,
    anchor=south,
    draw,
    inner sep=0.2em,
    draw
  ]at([yshift=1ex]legendpos)
  {	
	   ~\ref{plots:plot1}& {\small $\mathcal{T}=4$ $\mathcal{L}=1024$} &[5pt]
	   ~\ref{plots:plot2}& {\small $\mathcal{T}=2$ $\mathcal{L}=512$} &[5pt]\\
    	\ref{plots:plot3}& {\small $\mathcal{T}=1$ $\mathcal{L}=1$} &[5pt]
		\ref{plots:plot4}& {\small BH-SNE}&[5pt]\\
   };
\end{tikzpicture}

    \caption{\textbf{Computation time of the high-dimensional similarities} in the MNIST dataset by using BH-SNE~\cite{ref:bhtsne} and by using A-tSNE with different parameters. The obtained precision with different parameter settings is shown.}
    \label{fig:computation_time}
\end{figure}

The left chart in Fig.~\ref{fig:computation_time} shows the comparison of computation times, in logarithmic scale, of the high-dimensional similarities on the MNIST dataset obtained by our technique and by the BH-SNE algorithm. 
The right chart in Fig.~\ref{fig:computation_time} depicts the precision $\rho$ of the neighborhoods. 
The precision is given by Eq.~\ref{eq:precision} and it is computed using the exact and the approximated neighborhoods. 
Generally, our approach generates a good embedding very efficiently for any given dataset we tested. Fig.~\ref{fig:embeddings}(b-e) show the embeddings generated using the described parameter settings for the MNIST dataset after 1000 iterations. 
It can clearly be seen that we achieve comparable results more than two orders of magnitude faster than the BH-SNE implementation.

Fig.~\ref{fig:computation_time} shows how the precision decreases when increasing the data size for a fixed parameter setting.
The number of leaves (corresponding to data points) to visit, included in the parameter setting, is fixed indepentently of the data size. When the data size increases the same number of leaves, corresponding to a smaller fraction of the overall data, is visited, causing the lower precision.
Since these parameters are not exposed to the user, but rather only the precision value, this does not have an effect in the acutal use case.
In general, we can see that with a small reduction in precision, the computation time can be greatly reduced.

\begin{figure}[tb]
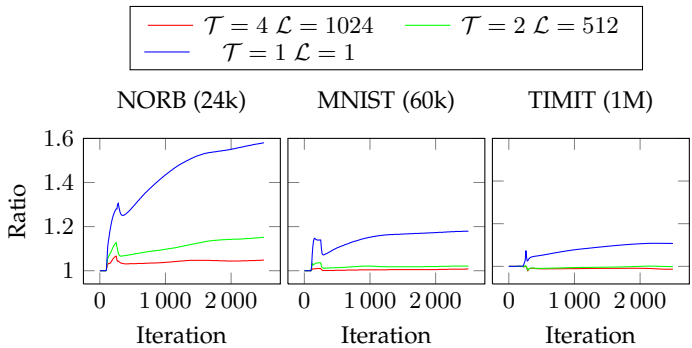


\begin{tikzpicture}

    \begin{groupplot}[group style={group size= 3 by 1,horizontal sep= 0.1cm},width=0.23\textwidth, height=3.5cm]
	    \nextgroupplot[title=NORB (24k),ylabel={Ratio},xlabel={Iteration},scaled x ticks = false, x tick label style={/pgf/number format/fixed,/pgf/number format/1000 sep = \thinspace},every axis/.append style={font=\small},legend style={font=\tiny},ymax = 1.6,]
			\input{charts/ratio_NORB.tex}
                \coordinate (top) at (rel axis cs:0,1);
\nextgroupplot[title=MNIST (60k),xlabel={Iteration},scaled x ticks = false, x tick label style={/pgf/number format/fixed,/pgf/number format/1000 sep = \thinspace},every axis/.append style={font=\small},legend style={font=\tiny},ymax = 1.6,yticklabels={,,}]
			\input{charts/ratio_MNIST.tex}
        \nextgroupplot[title=TIMIT (1M),xlabel={Iteration},scaled x ticks = false, x tick label style={/pgf/number format/fixed,/pgf/number format/1000 sep = \thinspace},every axis/.append style={font=\small},legend style={font=\tiny},ymax = 1.6,yticklabels={,,}]
			\input{charts/ratio_TIMIT.tex}
                \coordinate (bot) at (rel axis cs:1,0);
    \end{groupplot}
\path (top|-current bounding box.north)--
      coordinate(legendpos)
      (bot|-current bounding box.north);
\matrix[
    matrix of nodes,
    anchor=south,
    draw,
    inner sep=0.2em,
    draw
  ]at([yshift=1ex]legendpos)
  {	
	   ~\ref{plots:plot1}& {\small $\mathcal{T}=4$ $\mathcal{L}=1024$} &[5pt]
	   ~\ref{plots:plot2}& {\small $\mathcal{T}=2$ $\mathcal{L}=512$} &[5pt]\\
    	\ref{plots:plot3}& {\small $\mathcal{T}=1$ $\mathcal{L}=1$} \\
   };
\end{tikzpicture}

    \caption{\textbf{Approximated to exact cost ratio} on different datasets of increasing size. When the size of the data increases, the ratio of the approximated cost divided by the exact cost is reduced given the same set of parameters.}
    \label{fig:error_approximation}
\end{figure}

Finally, we analyze the error introduced by the approximation of the similarities in the high-dimensional space using the NORB, MNIST and TIMIT datasets.
For the results of the CIFAR-10 dataset we refer to the supplemental material, as they are very similar to the results obtained on the MNIST dataset.
The cost function $C(P,Q)$ is the most direct indication of the quality of the embedding and we compare minimizing of the cost function $C(P,Q^A)$ to $C(P,Q)$. 
$Q^A$ is the joint-probability distribution that describes similarities in the approximated embedding obtained by the minimization of $C(P^A,Q^A)$. 
Fig.~\ref{fig:error_approximation} shows the $C(P,Q^A)/C(P,Q)$ ratio.
Smaller values indicate less error, with a value of $1$ meaning that no approximation error is present.
The \emph{Early Exaggeration} of the attractive forces (see Sec.~\ref{sec:refinement}) is responsible for the peak in the ratio that is visible during the first 250 iterations. 
By exaggerating the attractive forces the approximation error is increased. 
Notice that the absolute value of the cost is not depicted in Fig.~\ref{fig:error_approximation} and decreases with every iteration.

The usage of a Forest of Randomized KD-Trees with $\mathcal{T}=1$ $\mathcal{L}=1$ generates an embedding with a large error. 
This configuration is an upper bound of the error and a lower bound in computation time: by visiting only one leaf during the traversal of the forest composed by just one tree, the approximated KNN algorithm becomes a \emph{greedy algorithm}~\cite{IntroductionAlgorithms}. 
We can also note that, when the size of the data increases, the approximation error decreases. 
In the TIMIT dataset we observe that the approximation errors generated using $\mathcal{T}=2$ $\mathcal{L}=512$ and $\mathcal{T}=4$ $\mathcal{L}=1024$, are similar or even better, than the exact one. 
By increasing the number of points in the embedding, the effect of the false repulsive forces (Sec.~\ref{sec:Our_tSNE}) is compensated by the increasing number of attractive forces among data-points.

The presented results clearly show that we can rapidly provide very accurate embeddings allowing immediate interaction, without misleading the user.
With a large number of data points we effectively generate tSNE embeddings as demonstrated by the reduced approximation error. 

\section{Interactive Analysis System}
\label{sec:interactive_system}

Using A-tSNE, the user is able to start the analysis of the data without waiting for the exact computation of the similarities in the high-dimensional space. 

The embedding, however, will be created based on approximated information.
We present different strategies that the user can apply to refine the high-dimensional similarities, leading to the generation of different, and more precise, embeddings.

During the refinement the user must be aware of the level approximation in the embedding.
Therefore, we present a visualization that supports the inspection of the level of approximation and, at the same time, is able to deal with the increased sized of the embeddings that A-tSNE is able to generate interactively.

We also take advantage of the steerability of A-tSNE to allow for a direct manipulation of the high-dimensional data and its representation, for example, by adding and removing data-points or by changing the high-dimensional representation of the data.  

Finally, we implemented these techniques in a coordinated multiple-views framework that allows for the direct inspection of the data in the embedding.

%
%
%

\subsection{User Driven Refinement}
\label{sec:refinement_strategies}
The refinement process used to steer the computation of an A-tSNE embedding works on a per-point basis, see Sec.~\ref{sec:refinement}.
We propose three different strategies that can be used to select the points to be refined.

The very basic strategy is to refine the neighborhoods of all the points in $X$ in a random order.
When computational resources are sparse, however, it makes sense to steer the refinement process to increase precision in areas of the embedding that the analyst finds interesting, e.g., based on initial visual clusters appearing in the embedding.

\subsubsection{User Selection}
We allow the user to select a subset of points for immediate refinement.
By brushing in the embedding the user can steer the refinement to interesting areas.
This strategy is less effective when just a few points are selected for refinement, as the forces exerted on its neighbors are still approximated, which can lead to an unfaithful description of the high-dimensional data.

\subsubsection{Breadth-First Search}
\label{sec:BreadthFirst}
If only a few points are selected for refinement, we extend the process to include their neighborhoods. Therefore, a breadth-first visit~\cite{IntroductionAlgorithms} of the graph created by the KNN relationships can be used to extend the refinement. If a priority queue~\cite{IntroductionAlgorithms} is used to keep track of the points that must be refined, different strategies can be implemented. For example, the priority in the queue can be given by the distances in the high-dimensional space or it can be determined based on a domain-specific criteria.

\subsubsection{Density-Based Refinement}
\label{sec:density_base_strategy}
If the user is more interested in gaining a global overview of the potential embedding, a density-based refinement strategy can be used instead of a local refinement. This strategy is based on the observation that points in the less dense areas of the high-dimensional space are responsible for the creation of the global relationship in a tSNE embedding~\cite{ref:tsne}. The data-points are refined with an order given by the density in the high-dimensional space. An indication of this density is the variance $\sigma_i$ of the Gaussian distribution, as explained in Sec.~\ref{sec:tSNE}. This strategy can work within a user-defined selection or on the whole dataset.

\subsection{Visualization and Interaction}
\label{sec:vis_embedding}

\begin{figure*}[t!]
        \centering
				  \subfloat[]{
						\includegraphics[width = 0.23 \linewidth]
						{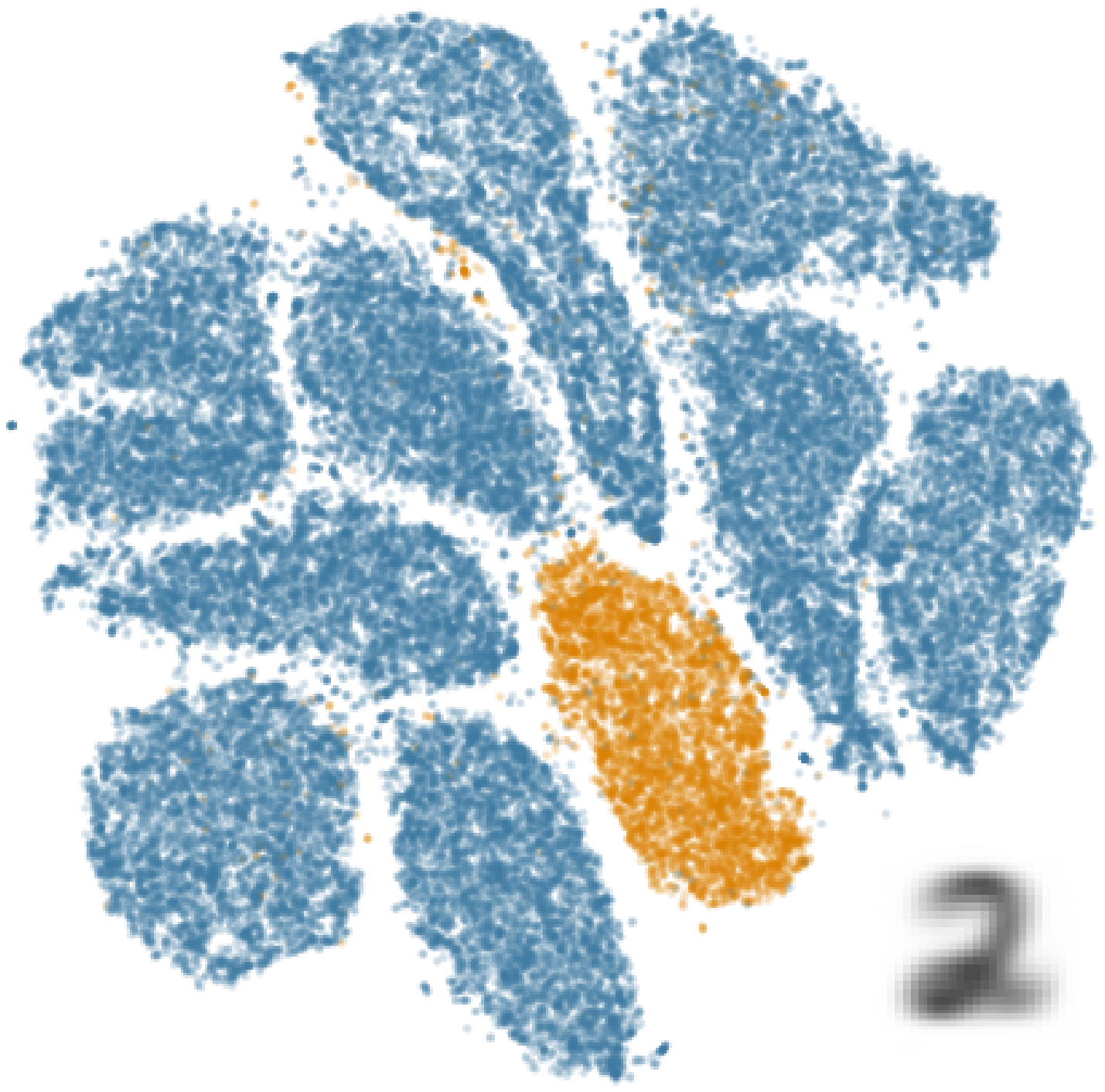}
						\label{fig:kde_visualization_0}
					}%
				  \subfloat[]{
						\includegraphics[width = 0.23 \linewidth]
						{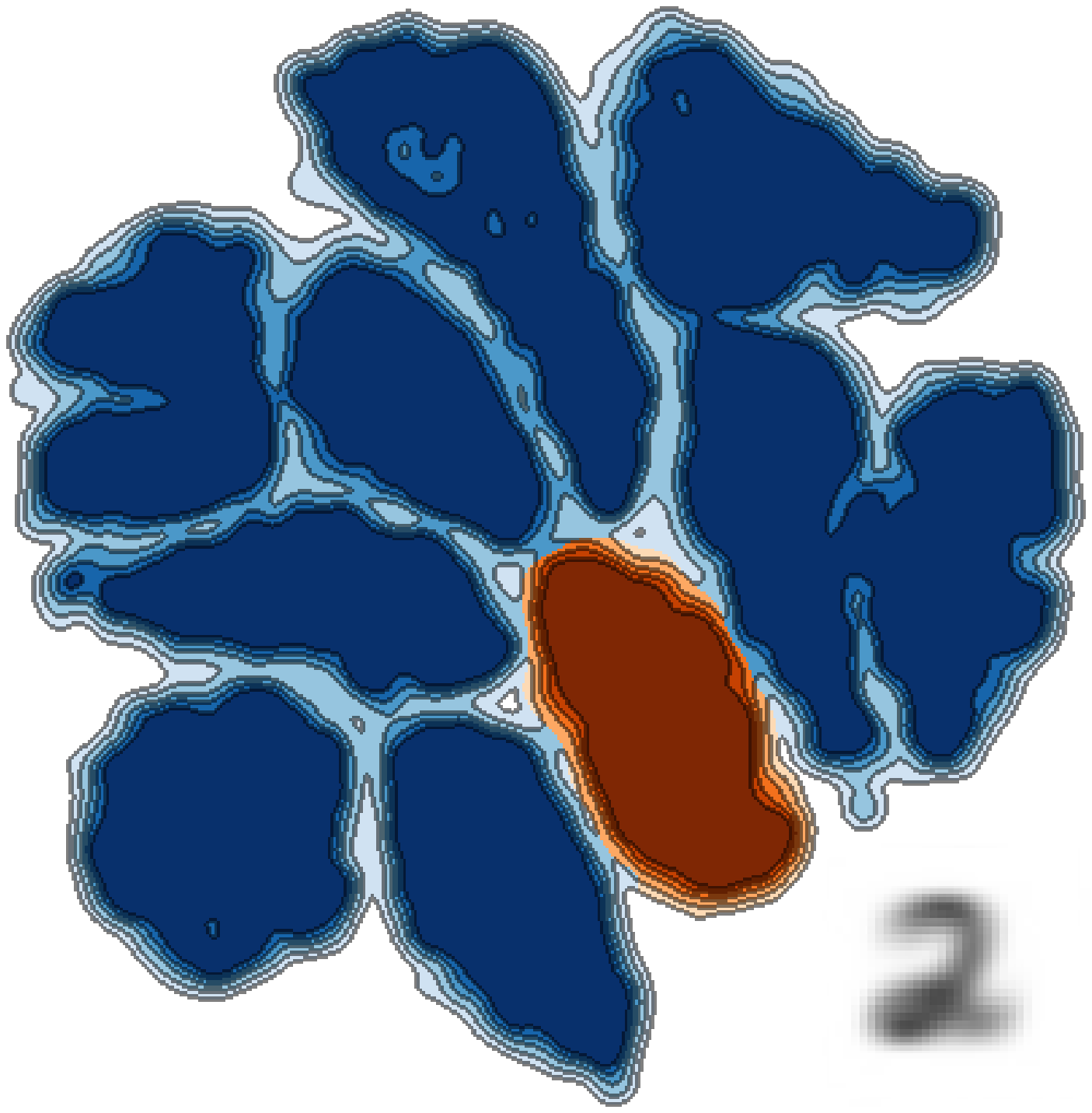}
						\label{fig:kde_visualization_1}
					}%
				  \subfloat[]{
						\includegraphics[width = 0.23 \linewidth]
						{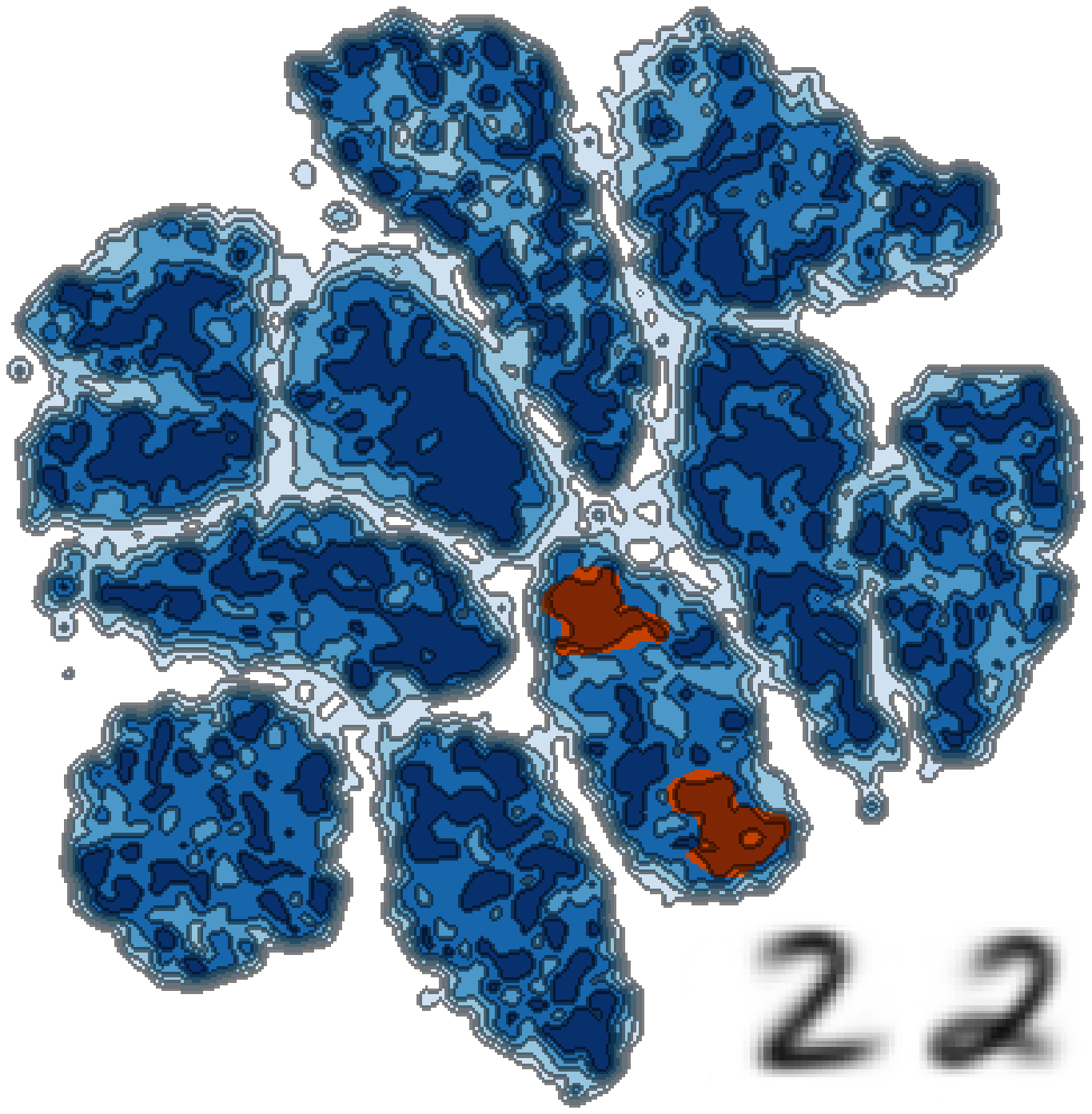}
						\label{fig:kde_visualization_2}
					}%
				  \subfloat[]{
						\includegraphics[width = 0.23 \linewidth]
						{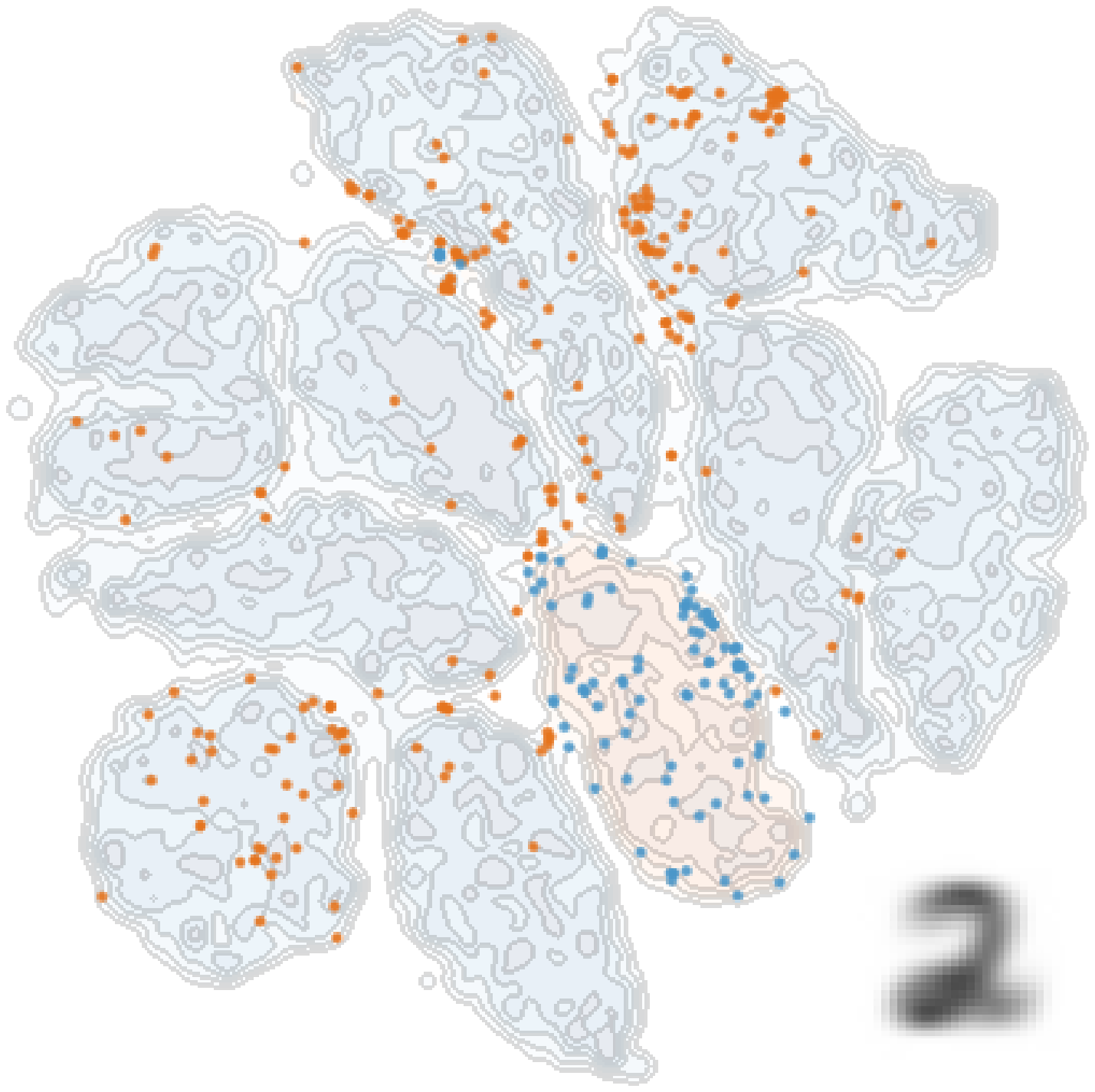}
						\label{fig:kde_visualization_outliers}
					}%
        \caption{\textbf{A-tSNE embedding of the MNIST dataset}. (a) uses a point-based visualization with an alpha value of $0.25$, the points colored in orange correspond to the digit '2'. (b,c) uses the real-time density-based visualization as described in Sec.~\ref{sec:kde}. By changing the bandwidth of the kernel density esitmation, clusters at different scales are visible. (d) shows the outliers in the data-points representing the digit '2' by means of a combination of the density-based and the point-based visualization. All figures show the average image of the selected clusters. }\label{fig:kde_visualization}
\end{figure*}

The visualization of the tSNE embedding is not insightful if not combined with the ability to inspect the high-dimensional data. 
Particularly in a context of exploratory data analysis, where a classification as the one presented in Fig.~\ref{fig:embeddings} is not available, such a solution is very helpful. 
In our system the user can explore the embedding using a density-based visualization. 
Selections in the embedding are used to visualize the high-dimensional data in a coordinated multiple-view framework. 
To indicate the approximation level in the embedding, we use two specifically-designed visualizations.

\subsubsection{Density-Based Visualization}
\label{sec:kde}

The visualization of the embedding, using simple points, is affected by visual clutter when the number of points increases. Density-based~\cite{silverman1986density} visualizations are commonly used to show a tSNE embedding~\cite{visne,becher2014high,ACCENSE,LastFromLaurens} because of their ability to visualize features at different scales. We apply a real-time kernel density estimation (KDE)~\cite{ref:KDE_Scatter_plots} for the creation of an interactive density-based visualization of the embedding.
We use changes in the color hue to visualize selections, for example to highlight data points that are selected to be analyzed in other views of the coordinated multiple-view framework.
The KDE is computed by assigning a value for each pixel $\mathbf{p}$ using the \emph{kernel density estimator} $f(\mathbf{p},h)$  as follows:

\begin{equation}
\label{eq:kernel_density_estimator}
 f(\mathbf{p},h) = \frac{1}{N} \sum^N_{i=1}G(||\mathbf{p}-\mathbf{y}_i||,h).
 \end{equation}
%

$G(d,h)$ is a zero mean Gaussian distribution with standard deviation $h$, which can be interactively chosen by the user in order to reveal clusters at different scales. 
Additionally, we introduce a transfer function, mapping $f(\mathbf{p},h)$ to a color, in order to highlight user-defined selections. 
Areas with a large percentage of selected points are visualized with a different transfer function, and selection outliers are shown as points. 
To achieve this goal, we introduce a new kernel density estimator $s(\mathbf{p},h)$, which illustrates the density of the user selection in a pixel $\mathbf{p}$. Given a set of selected data-points $S$ we use:

\begin{equation}
\label{eq:kernel_density_estimator_selection}
s(\mathbf{p},h) =\frac{1}{f(\mathbf{p},h)} \frac{1}{|S|} \sum_{\mathbf{y}_i \in S} G(||\mathbf{p}-\mathbf{y}_i||,h)
\end{equation}

If $s(\mathbf{p},h)$ is higher than a threshold $S_{thresh}$, a transfer function based on a different hue and with a higher luminance is used. 
We found empirically that a value $S_{thresh} = 0.5$ performs satisfactorily without compromising the quality of the visualization. 
We also use a point-based visualization of isolated selected data-points and, unselected data-points in selected regions.  
Finally, the user has control over the opacity of points or opacity of the density-based visualization to adjust the visualization to the needs of the analysis. 
An example of different visualizations of the embedding is presented in Fig.~\ref{fig:kde_visualization} where the MNIST dataset is used. 
The analyst can change the bandwidth $h$, the transfer function, and the opacity interactively in order to show clusters at different scales and outliers in the selection.

For example, Fig.~\ref{fig:kde_visualization_1} shows the selection of a high-level cluster.
If a different bandwith is chosen, as in Fig.~\ref{fig:kde_visualization_2}, clusters at a different level appear.
Finally, if the labels are used to make a selection in the embedding, as in Fig.~\ref{fig:kde_visualization_outliers}, it is possible to see the distribution of the outliers in the density-based visualization.

\subsubsection{Visualization of the Approximation}

During the refinement, the precision of the high-dimensional similarities is gradually refined until exact.
In order to give to the user the ability to inspect the level of approximation, we enhance our density-based visualization to show the requested precision $\rho_i$.
Note that $\rho_i$ is different for every data-point and changes during the refinement process, as described in Sec.~\ref{sec:refinement}


For each pixel $\mathbf{p}$ we assign a value given by the function $a(\mathbf{p},h)$ that represents the approximation value given the bandwidth $h$:

$$ a(\mathbf{p},h) = \frac{1}{f(\mathbf{p},h)} \frac{1}{\sum^N_{i=1}\rho_i} \sum^N_{i=1} \rho_i G(||\mathbf{p}-\mathbf{y}_i||,h) $$

$a(\mathbf{p},h)$ is the precision $\rho_i$ weighted kernel-density divided by the kernel-density estimator $f(\mathbf{p},h)$.
The value $a(\mathbf{p},h)$ is between zero and one and can be used directly for encoding of the approximation in the visualization.

The value of the function $a(\mathbf{p},h)$ can then be included in the visualization in two different ways.
First, we introduce a Magic Lens~\cite{MagicLenseSurvey} that shows the approximation with a minimal conceal of the data.
We use a circular lens that can be overlayed on the density-based visualization and $a(\mathbf{p},h)$ is used to define the transparency $\alpha$ of every pixel in the lens. To better highlight the refined areas, we use $\alpha = 1-a(\mathbf{p},h)^k$, where $k$ is a user selected parameter, to compute $\alpha$. 
We provide a default value of $k=2$. 

\begin{figure}[t]
        \centering
				  \subfloat[Magic Lens]{
						\includegraphics[width = 0.49 \linewidth]
						{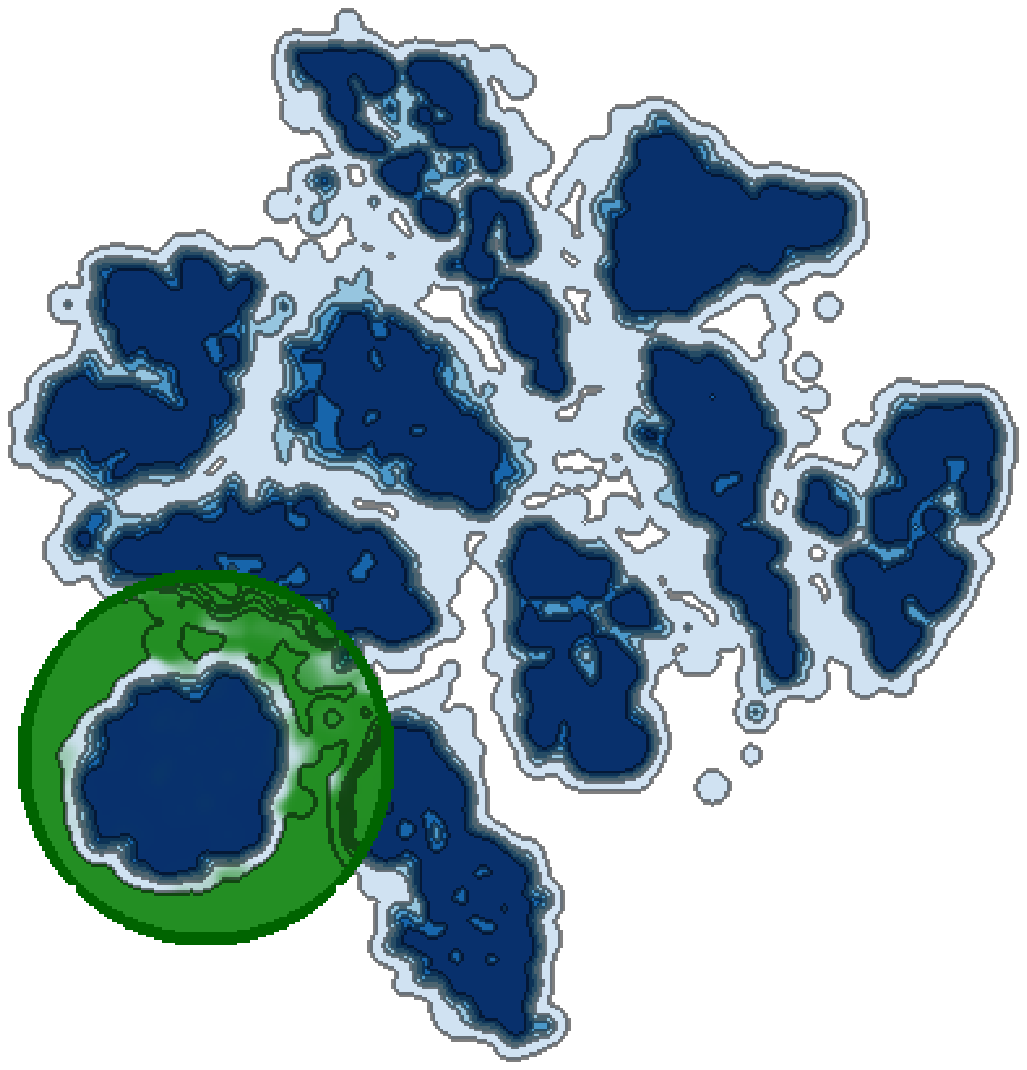}
						\label{fig:approximation_lens}
					}%
				  \subfloat[Full View Mode]{
						\includegraphics[width = 0.49 \linewidth]
						{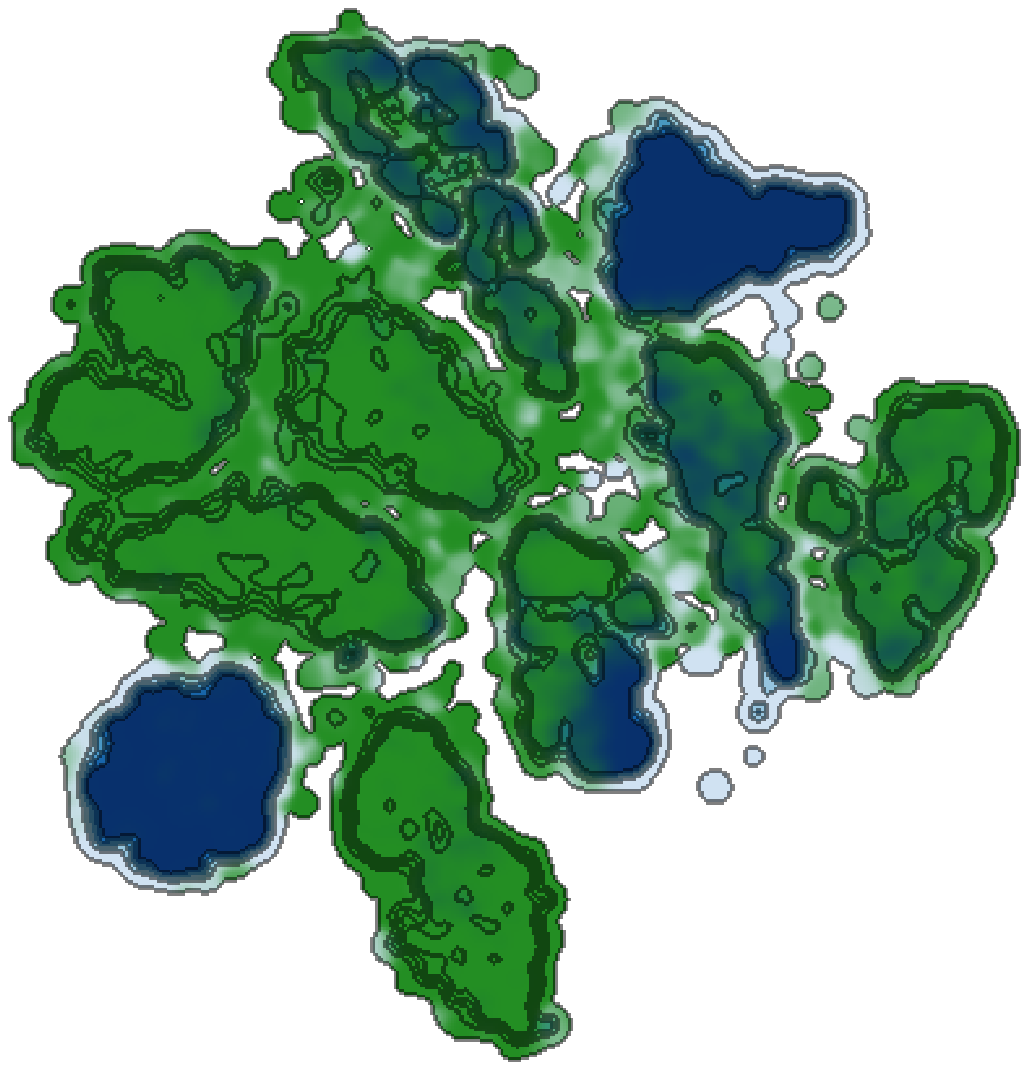}
						\label{fig:approximation_mode}
					}%
        \caption{\textbf{Visualization of the approximation} in the embedding by means of a magic lens (a) and the full view mode (b).}
				\label{fig:approximation}
\end{figure}

Fig.~\ref{fig:approximation_lens} shows the lens over a cluster that is already refined and, therefore, is visible through the lens.
The green tone indicates the area where similarities are still approximated.
Contours in approximated areas are preserved to indicate the structure of the embedding.
We color the areas without points in green to put more emphasis on refined areas.

In addition to the Magic Lens, we provide the possibility to map approximation to the complete view.
This is especially useful to inspect the embedding when one of the global refinement strategies was selected.
Fig.~\ref{fig:approximation_mode} shows the approximation in the embedding using this approach. 
It is possible to see that two clusters are already refined, relying on exact neighborhood relationships.
The user selected a \textit{Breadth-first search} refinement strategy, therefore, the refinement is spreading through the embedding, leading to some areas in the top-right corner having the original color.
However the perception of clusters is reduced by removing the color information inside the contours.

\subsection{Data Manipulation}
\label{sec:data_manipulation}
In Sec.~\ref{sec:refinement}, we show that we are able to update high-dimensional similarities between data-points during the gradient-descent minimization.
In this section, we take advantage of this possibility, introducing different operations that the analyst can use to manipulate the original data-points in their high-dimensional feature space.

The embedding does not need to be recomputed but evolves dynamically as the data changes. 
At the center of an interactive exploration of data is the ability to add or remove data on demand, use different representations of the same dataset or adapt to any changes in the data.

\subsubsection{Inserting Points}
For a point $\mathbf{x}_a$, which we want to add to the embedding, its neighborhood $\mathcal{N}_a$ needs to be computed. An approximated algorithm can be used to compute the neighborhood and a refinement can be scheduled. To complete the insertion we need to check whether $\mathbf{x}_a$ belongs to the KNN of each point in $X$. We define $d_i^\text{Max}$ as the maximum distance between a point $\mathbf{x}_i$ and the points in its neighborhood $\mathcal{N}_i$. The update of the neighborhoods is written as follows: 

 \begin{equation}\label{eq:pareto mle2}
  \begin{multlined}
\forall \mathbf{x}_i \in X \text{   if  } || \mathbf{x}_a - \mathbf{x}_i || < d_i^\text{Max} \\
 \text{  then  } \mathbf{x}_a \in \mathcal{N}_i \text{ and } \mathbf{x}_j \not \in \mathcal{N}_i : || \mathbf{x}_i - \mathbf{x}_j || = d_i^\text{Max}
 \end{multlined}
\end{equation}

This operation is computed in $O(N)$ if $d_i^\text{Max}$ is cached. A priority queue is used to efficiently update $d_i^\text{Max}$ after the insertion of $\mathbf{x}_a$ in a given neighborhood $N_i$. It is important to observe that the insertion of $\mathbf{x}_a$ in $\mathcal{N}_i$ will not reduce the estimated precision $\rho_i$. The initial position in the embedding $\mathbf{y}_a$ is given by the average position of its neighbors $\mathcal{N}_a$ weighted by their similarity $p_{j|i}: \mathbf{x}_j\in \mathcal{N}_i$.
The new point $\mathbf{x}_a$ must be added in the \textit{Forest of Randomized KD-Trees}. This operation can be performed in $O(\log(N))$ .

\subsubsection{Deleting Points}

Removing a point $\mathbf{x}_r \in X$ is performed by deleting $\mathbf{x}_r$ from the KNN of every point $\mathbf{x}_i \in X$. This operation has a computational complexity of $O(N)$. 
By removing $\mathbf{x}_r$ from a neighborhood $\mathcal{N}_i$ we reduce the number of $\mathbf{x}_i$ neighbors to $K-1$ and a new neighbor must be found to maintain the precision level. 
However, the new point in the neighborhood is the most dissimilar of the points in $\mathcal{N}_i$ thus its attractive force is rather small and we propose to ignore the contribution of the missing point, decreasing the estimated precision $\rho_i$ by $1/K$.
As for the insertion of a new data-point, the \textit{Forest of Randomized KD-Trees} is updated in $O(\log(N))$.



\subsubsection{Dimensionality Modification}

We handle changes in the description of a single high-dimensional data-point, for example, when the data is time varying, by a combination of removal and addition operations.
If the user wants to completely change the high-dimensional representation of the data, e.g., by adding or removing dimensions, a new approximated joint-probability distribution $P^A$ is computed and the embedding is free to evolve accordingly.
\subsection{Visual Analysis Tool}

\begin{figure}[!t]
\centering
        \includegraphics[width=0.48\textwidth]{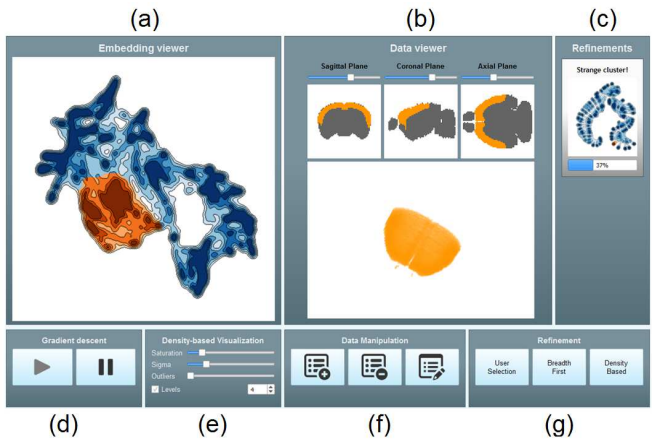}
        \caption{\textbf{Screenshot of our integrated system} using multiple linked views for interaction. The system comprises an embedding viewer (a), a data viewer (b) and a refinement viewer (c). Controls on the gradient descent (d), the density-based visualization (e), the data-manipulation (f) and the refinements (g) are at the bottom of the interface. }\label{fig:system}
\end{figure}

We implemented A-tSNE as a module in an integrated, interactive, multi-view system for the analysis of high-dimensional data.
Fig.~\ref{fig:system} shows a screenshot of the system and its different views.

The interface is divided into two main areas.
At the top, three different views are used to show the intermediate embeddings (7a), the data (7b) and the state of refinement processes (7c), respectively.
Controls on the generation of intermediate embeddings (7d), visualization of the embedding (7e), data manipulation (7f) and refinement (7g) are at the bottom of the interface.


The data subject to the analysis are visualized in the \emph{Data View} (7b).
Selections in the embeddings can be reflected in the Data View with strategies that depend on the data type.
For example, in Fig.~\ref{fig:system} the Data View is used to visualize data-points that represent voxels in a 3D volume viewer and using slices of the volume along the three main axes (7b).
Selections are highlighted in the anatomical planes by a change of hue and by adjusting the transfer function in the volume viewer.
The Data View can support selections based on a data-driven criteria, e.g. voxels with the same axial coordinate. 

We implemented multiple widgets that can be switched out in the Data View, to support the analysis process of different data types.
These widgets include a heatmap view and an image view.
If necessary multiple and different views can be combined for the analysis.

The \emph{Refinement-Status View} (7c) is used to give an overview of the progress of the refinements triggered by the user.
The user can steer the evolution of the embedding by refining areas with strategies as described in Sec.~\ref{sec:refinement_strategies}.
A refinement process is identified by the snapshot of the embedding when the user triggered the refinement, a user-defined description, and a progress bar that shows the percentage of the refined data-points over the selected ones.

\section{Implementation}

We implemented the system using a combination of C++ and Qt, as well as OpenGL with custom shaders in GLSL for the visualization of the embedding.
Where possible, we exploited parallelizability of our approach. 
The approximated neighborhoods are computed using the FLANN library~\cite{muja_flann_2009}, which implements KNN algorithms.
The density-based visualization is computed on the GPU using OpenGL and GLSL shaders. 
A precomputed floating-point texture is generated using a Gaussian kernel. 
A geometry shader is used to generate a quad for each point that is colored using the precomputed texture, the KDE is obtained by drawing into a Frame Buffer Object using an additive blending~\cite{ref:KDE_Scatter_plots}. 

\begin{figure*}[tb!]
    \centering
	\includegraphics[width=\textwidth]{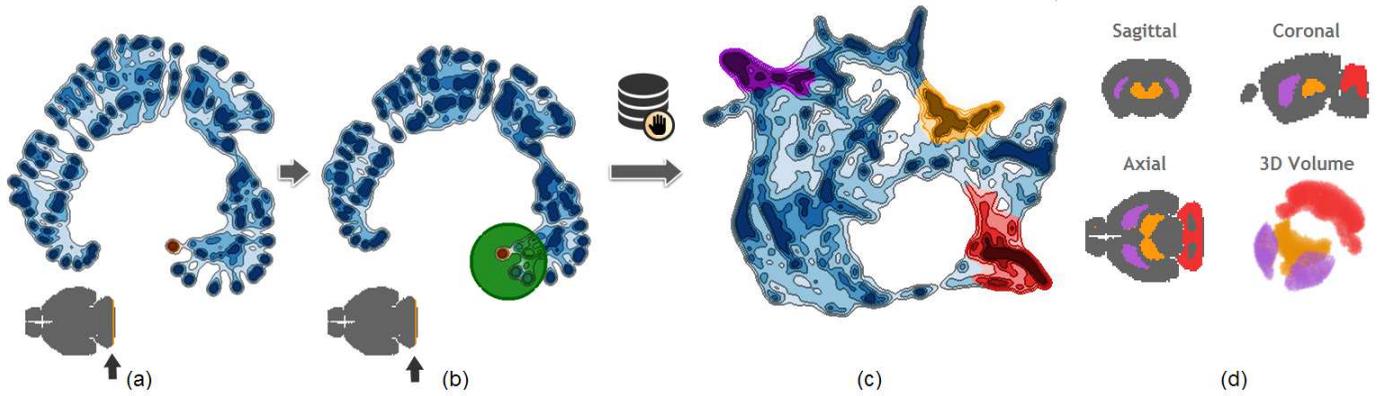}
    \caption{\textbf{Analysis of the gene expression in the mouse brain using A-tSNE}. The first embedding (a) are generated in $\approx 51$ seconds while 3 hours and 50 minutes are required by BH-SNE. The analyst inspects a cluster and finds that it corresponds to a slice in the data. The cluster do not disappear after the neighborhoods are refined, as shown by the lens in (b). A change in the high-dimensional data reveal that genetic information can be used to differentiate anatomical regions. (c) shows the final embedding based on a small number of Principal Components where three clusters are highlighted and (d) shows the corresponding regions in the brain.}
	\label{fig:aba_workflow}
\end{figure*}

\section{Case Study I: Exploratory Analysis of Gene Expression in the Mouse Brain}
\label{sec:aba}
%
%
%

In this section, we demonstrate the advantages of using A-tSNE in our visual analysis tool for the visual analysis of high-dimensional data.  Therefore, we present a use case, based on the work by Mahfouz et al.~\cite{mahfouz2014visualizing}, who use tSNE to explore the Allen Mouse Brain dataset~\cite{MouseBrainAtlas}.
The dataset is composed by 61164 voxels obtained by slicing the mouse brain in 68 slices. 
Each voxel is a 4345-dimensional vector, containing the genetic expression at the corresponding spatial position.
tSNE is computed using the voxels as data-points and the expression of the genes as high-dimensional space.
Note that no spatial information is used to build the high-dimensional space.

Mahfouz et al. discuss the hypothesis that genetic information can be used to differentiate anatomical structures in the brain.
Some regions in the brain, e.g. the Cerebellum, are known to have a highly different genetic footprint compared to the rest of the brain. 
In their work, Mahfouz et al. demonstrated that tSNE is effective in separating different anatomical structures, e.g. white and grey matter, based on their genetic footprint.

Fig.~\ref{fig:aba_workflow} depicts the typical analytic workflow using our visual analysis tool.
The first goal during the analysis is to validate the input data.
The acquisition process may not be perfect, data can be incomplete or noisy, therefore, it must be re-acquired or preprocessed before interesting results can be generated.
Driven by the need to validate the data as soon as possible, the user selects a reasonably low value for the desired precision, e.g. $\rho=0.2$, that will be used to estimate the parameters of the KNN algorithm.
With such a parameterization, A-tSNE computes the high-dimensional similarities in $\approx 51$ seconds while 3 hours and 50 minutes are required by BH-SNE.

The user then start analyzes the intermediate embeddings, produced by A-tSNE, in order to validate the input data.
After $\approx 170$ seconds several clusters become visible in the embedding as depicted in Fig.~\ref{fig:aba_workflow}a. 
The clusters are stable for several iterations indicating that they are not an artifact of the minimization process but represent clusters in the high-dimensional space.
The user can validate this by selecting the clusters in the embedding and inspect them in more detail, for example, by highlighting their spatial positions in the feature view, see Fig~\ref{fig:aba_workflow}a.
Points or clusters are selected by brushing in the embedding. 
During a brushing operation the generation of intermediate embeddings is stopped to make sure the user does not accidentally brush areas as they change.
Selected points are then highlighted by a change of hue, in this case from blue to orange.
Further inspection using the Data View in our interactive system, shows that each cluster corresponds to a slice in the dataset.
Fig.~\ref{fig:aba_workflow}a shows a cluster, highlighted in orange, and the corresponding slice in the volume.

To make sure the clusters are not an artifact introduced by the approximated similarities, the user can refine the selected data-points while the embedding evolves.
Fig.~\ref{fig:aba_workflow}b shows the embedding after the refinement is complete.
The user can inspect the degree of approximation in the embedding using the interactive lens.
The lens is less transparent over approximated areas of the embedding and transparent on the areas that contain no approximation, i.e., the selected points.
After the refinement of the high-dimensional similarities of the selected data points, the clusters do not disappear and it becomes clear that the clusters are caused by properties of the data, rather than by the approximation.

Therefore, the user can stop the generation of the embedding.
Further analysis of the input data reveals that missing values in the input data cause the formation of small clusters in the embedding.
Mahfouz et al. removed this effect by using the first $10$ components, extracted by a Principal Component Analysis of the raw data, as the high-dimensional space.

In our system, the user can now directly change the high-dimensional space, without restarting the computation of the embedding.
Approximately $200$ seconds after the change in the high-dimensional data, a stable embedding is obtained. 
Fig.~\ref{fig:aba_workflow}c shows the final embedding, where three different clusters are highlighted.
Fig.~\ref{fig:aba_workflow}d depicts the selected voxels in the brain, note how the anatomical structures are now revealed.
It is possible to see how the clusters that were present in the first intermediate results disappear, showing that the cluster fragmentation is removed.

Voxels that belongs to the same anatomical structure are close together in the embedding.
A-tSNE is able to separate anatomical structures based on the gene expression of the 4345 genes.

In their work, Mahfouz et al., presented embeddings created using 2, 3, 5, 10 and 20 principal components as the high-dimensional space.
Identifying the right number of components is a time consuming task and the adoption of our analytic workflow helps the user in finding a good compromise interactively analyzing the resulting embedding generated changing the number of components.

\section{Case Study II: Real-time monitoring of high-dimensional streams}

\begin{figure*}[t]
  \centering
  \subfloat[]{
    \includegraphics[width = 0.25 \linewidth]
    {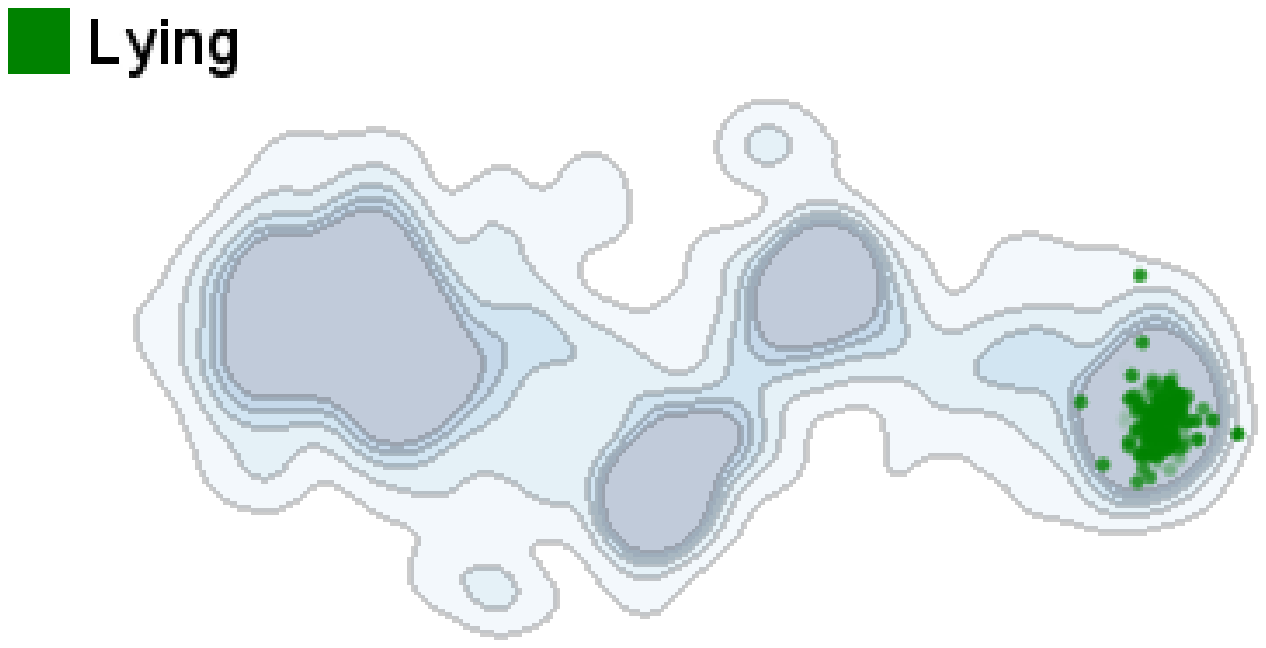}
    \label{fig:real_time_streaming_0}
  }%
  \subfloat[]{
    \includegraphics[width = 0.25 \linewidth]
    {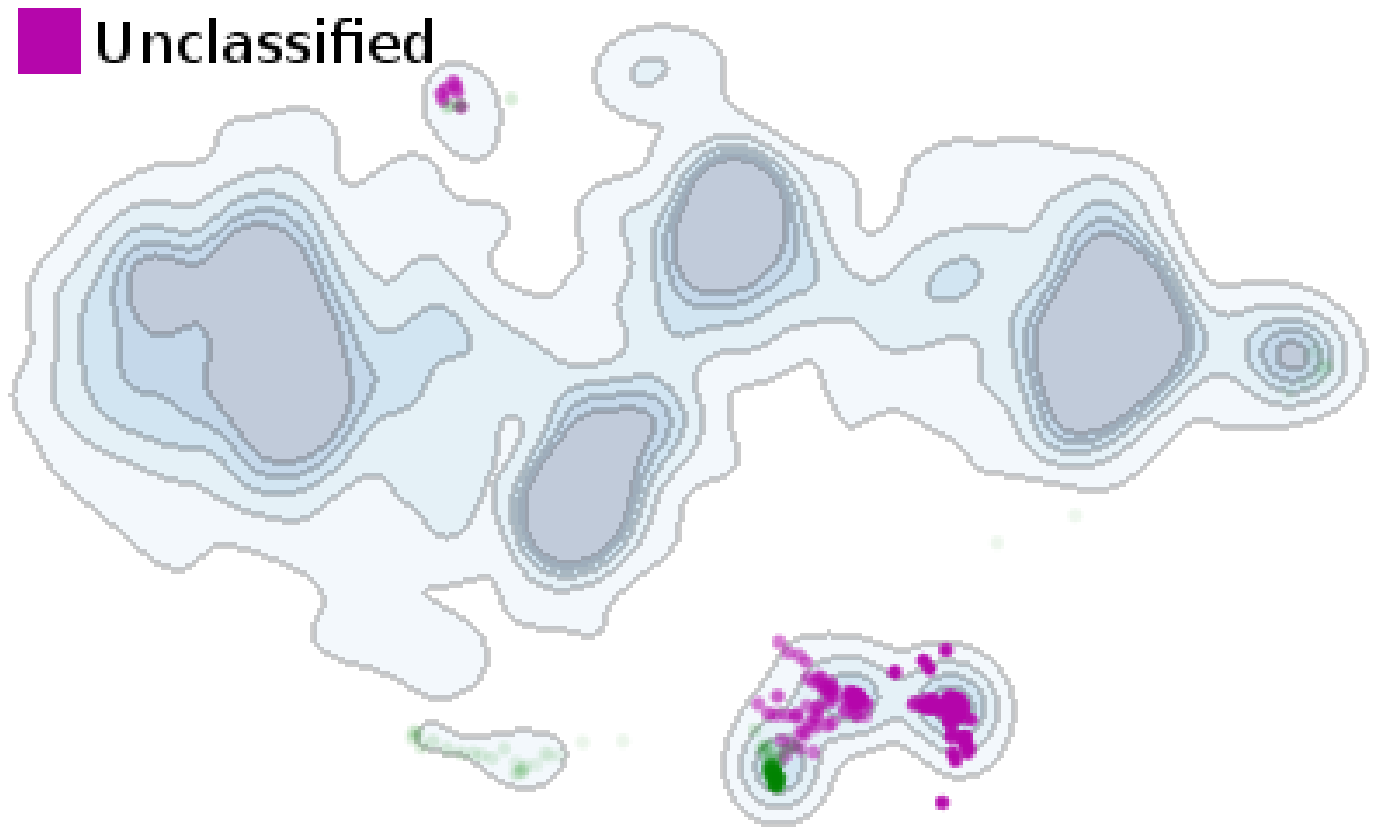}
    \label{fig:real_time_streaming_1}
  }%
  \subfloat[]{
    \includegraphics[width = 0.25 \linewidth]
    {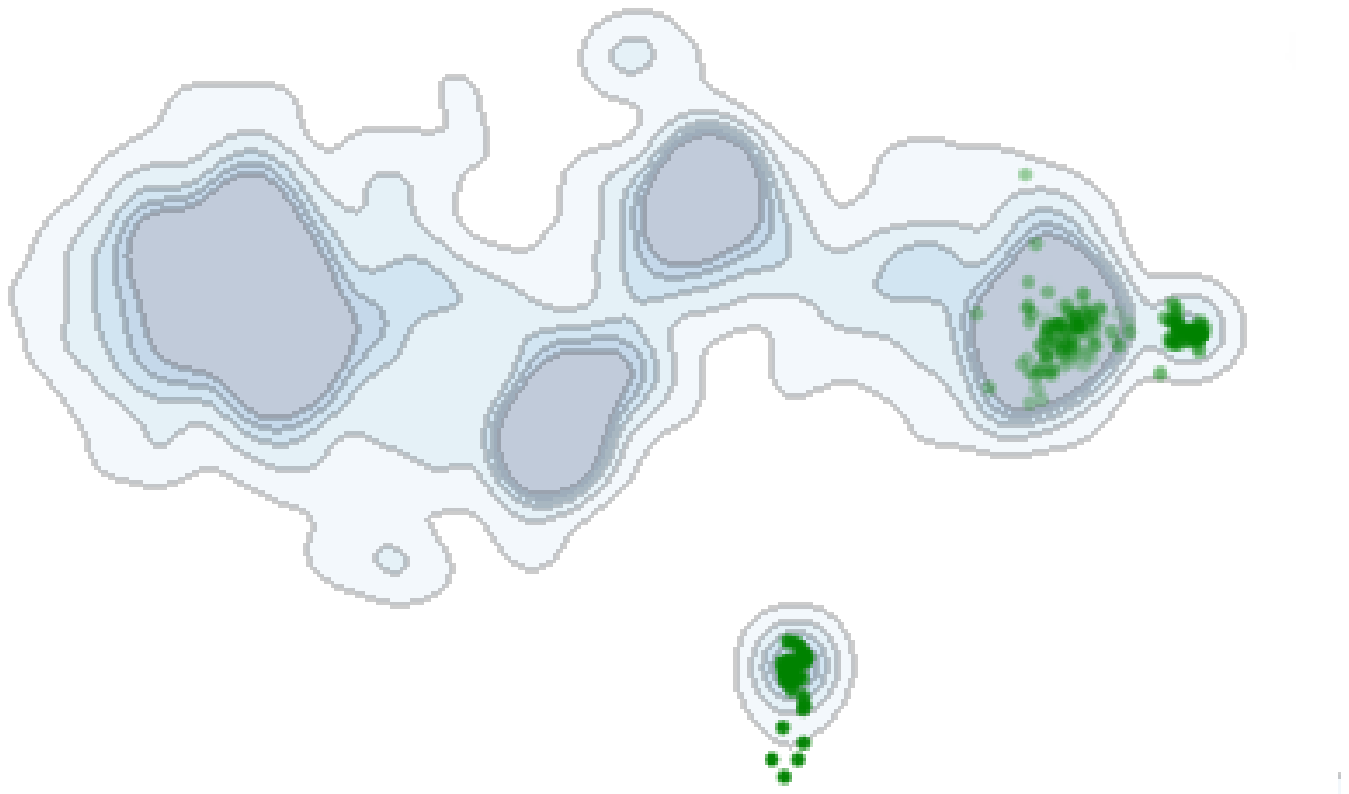}
    \label{fig:real_time_streaming_2}
  }%
  \subfloat[]{
    \includegraphics[width = 0.25 \linewidth]
    {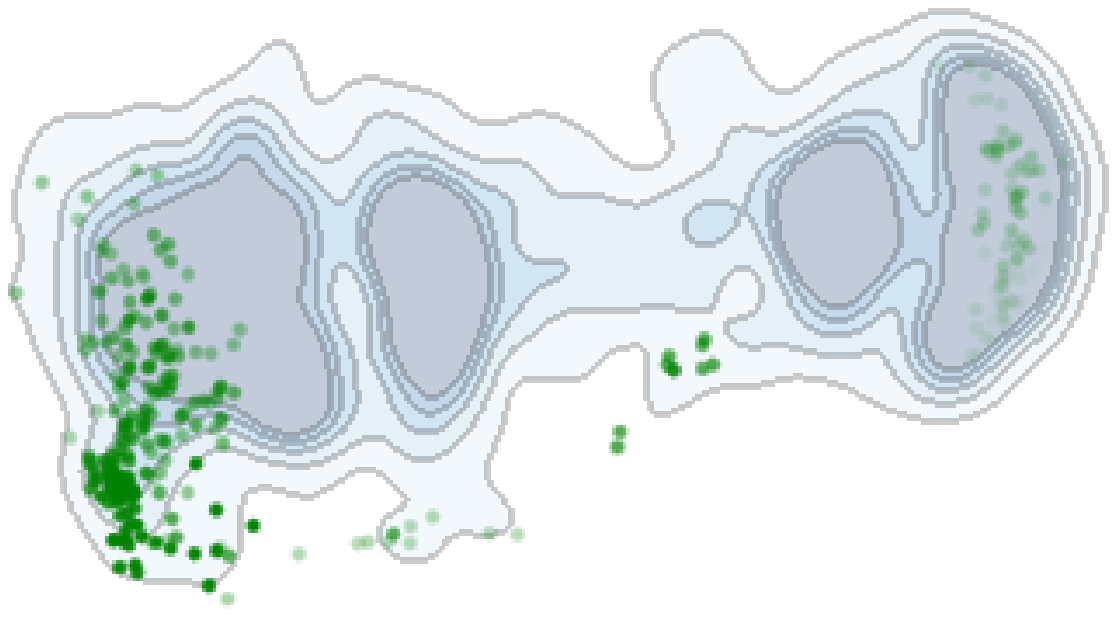}
    \label{fig:real_time_streaming_3}
  }
\caption{\textbf{A-tSNE is used for the real-time analysis of high-dimensional streams}. The embeddings are generated using the readings of the last 10 minutes. As new readings arrive they are inserted in the embedding and they are highlighted using a point-based visualization. (a) shows the initial embedding, the color of the data-points indicates that the subject is lying down.
The embedding evolves as in (b), the new cluster indicates the creation of a set of different readings. This insight is confirmed by a change in the color of the data-points that indicates a new type of label activity.
(c) shows an evolution of the embedding presented in (a) where new readings are generated from a miscalibrated sensor and, therefore, are clustered together.
By removing the features corresponding to the miscalibrated sensor the embedding evolves as in (d). The cluster that identifies miscalibrated readings is removed.
}
  \label{fig:real_time_streaming}
\end{figure*}%

The improved computation time and the ability to manipulate data are the key for applying tSNE in new application scenarios, such as the real-time monitoring of high-dimensional data streams.
The original tSNE algorithm fails in providing a solution for such an application.
The computation of a tSNE map imposes a time constraint that cannot be ignored, when the rate in which new data is generated is higher than the time required for the computation of a tSNE map.

As proof of concept, we selected a dataset for physical activity monitoring~\cite{IMUsDataset} that comprises readings of three Inertial Measurement Units (IMU) and a heart rate monitor applied to 9 different subjects.
Every IMU generates 17 readings every 10 ms, while the heart rate monitor generates one reading every 100 ms.
Taking into account all the sensors, we have a stream of data consisting of 52 readings, where a new data is generated every 100 ms for each subject.
Every subject has also a device to label the physical activity among 24 different activities, e.g. lying down, standing, running.
We use the labeling of every reading to validate the insights obtained by the analysis of the embeddings.

We analyze the stream of a subject by keeping the readings of the previous $\mathcal{M}$ minutes in the embedding.
When a new reading is generated, we add it to the embedding using the technique described in Sec.~\ref{sec:data_manipulation}. 
Similarly, when a reading is older then $\mathcal{M}$ minutes, we remove it from the embedding.
In the test presented in this section, $\mathcal{M} = 10$ is set leading to an embedding composed, in average, by 6000 data-points that is updated every 100 ms.

We add a point-based visualization to our density-based visualization, which shows the last points inserted in the embedding.
The new points are colored according to the classification of the activity made by the subject and they will fade out in $\mathcal{F}$ seconds. 
By showing the new data-points the analyst can identify where new points are added, providing at the same time an overview of the embedding in the last $\mathcal{M}$ minutes and the trend of the last $\mathcal{F}$ seconds.

Fig.~\ref{fig:real_time_streaming}a shows an embedding obtained from \textit{subject 105}, where the color of the data-points, green in this specific case, indicates that the subject is lying down.
The embedding is composed of a single big cluster that represent the \textit{lying down activity}.
The cluster is divided in four different sub-clusters that identify different readings of the sensors.
The readings of the last 30 seconds belong to a single sub-cluster and can be seen as points on the right side of the embedding.
The embedding evolves based on new readings from the sensors, after few seconds the new data-points start to be placed further away from the original cluster, leading to the creation of a new cluster, as depicted in Fig.~\ref{fig:real_time_streaming}b.
After a few seconds the subject changes the classification of his activity from lying down to an \textit{unclassified activity}, whose corresponding data-points are colored in purple.
It is interesting to note that, simply by looking at the embedding, it is possible to predict a change in the labeled activity before the subject is able to record the change on his labeling device.
It can be seen by the fact that few data-points labeled as a \textit{lying down activity}, hence colored in green, are in the same cluster as the ones identified as \textit{unclassified activity}.
In this particular case, we can guess that the subject seated before changing the labeled activity.

Finally, we simulated a miscalibration in an inertial measurement unit.
Differently from a faulty sensor (not generating any readings), a miscalibrated one generates readings affected by a constant offset that is different for every dimension.
We simulate the miscalibration by enforcing a random offset to the readings generated by one of the IMUs.
A miscalibrated sensor generates readings that are different from the normal one and, therefore, they should be clustered together as faulty readings. 
Fig.~\ref{fig:real_time_streaming}c shows the evolution of the embedding presented in Fig.~\ref{fig:real_time_streaming}a where the miscalibrated readings are grouped by A-tSNE.
After the inspection of the readings generated from the IMUs, the analyst can identify that something is wrong with one of the sensors.
At this point the sensor may be replaced or, in case this is not possible, the readings from the miscalibrated sensor can be excluded by removing the corresponding dimensions from the high-dimensional space.
Fig.~\ref{fig:real_time_streaming}d shows how the previous embedding evolves when the readings generated by the miscalibrated sensor are removed from the high-dimensional space.
It is possible to see that the readings affected by the miscalibration are now close to the cluster that represent the \textit{lying down activity}.

To conclude, we demonstrated how A-tSNE can be used for the real-time analysis of high-dimensional streams.
It is not possible to compute a standard tSNE embedding every 100 ms without the fast computation of the high-dimensional similarities and the ability to directly manipulate the high-dimensional data.

\section{Conclusions}
Motivated by the need of interactivity in Visual Analytics, we developed Approximated-tSNE (A-tSNE). A-tSNE enables the rapid generation of approximate tSNE embeddings.
We used a fast approximated KNN algorithm for the computation of the high-dimensional similarities.
Our algorithm is designed to be used within the Progressive Visual Analytics context, allowing the user to have a quick preview of the data.
The insight obtained using the approximated embeddings can be validated by removing the approximation in interesting areas with different strategies.
The user is made aware of the level of approximation in the embeddings by the usage of two specifically designed visualizations.

We demonstrate that A-tSNE generates meaningful tSNE embeddings two orders of magnitude faster than the state-of-the-art, BH-SNE.
Furthermore, our method enables a new way of handling large datasets, by integrating a real-time density-based visualization to support the user in interacting with the data.
The selections can be visualized in a coordinated multiple-view framework making it easy to steer the analysis into relevant regions.
A-tSNE can effectively replace BH-SNE for all use cases, as the full precision of BH-SNE can always be reached by setting the precision parameter accordingly, or refining the data.
We present three different operations for the direct manipulation of the high-dimensional data and their application for the real-time monitoring of high-dimensional streams.

In our work we focus on the fast computation of high-dimensional similarities enabling interactive analysis. However, when dealing with Big Data, e.g., more than a million data-points, the iterative minimization becomes slower and limits the interactivity of A-tSNE.
In the future we want to address this limitation and explore the application of A-tSNE in other research scenarios. In particular, we are interested in investigating the application of A-tSNE in the analysis of heterogeneous data and different high-dimensional streams, such as climate readings.

\ifCLASSOPTIONcaptionsoff
  \newpage
\fi



%
%
%

\bibliographystyle{IEEEtran}
\bibliography{bibliography}
\end{document}